\pdfoutput=1
\documentclass[11pt]{article}

\usepackage[final]{acl} 

\usepackage{times}
\usepackage{latexsym}
\usepackage[T1]{fontenc}
\usepackage[utf8]{inputenc}
\usepackage{microtype}
\usepackage[scaled=1.0]{zi4}
\usepackage{graphicx}
\usepackage{makecell}
\usepackage{booktabs}
\usepackage{adjustbox}
\usepackage{float}
\usepackage{placeins}
\usepackage{stfloats}
\usepackage{multirow}

\usepackage{tcolorbox}
\usepackage{setspace}
\usepackage{array}[=2016-10-06]

\title{Finding Diamonds in Conversation Haystacks: \\A Benchmark for Conversational Data Retrieval}

\author{
    \textbf{Yohan Lee\textsuperscript{1,2}$^*$},
    \textbf{Yongwoo Song\textsuperscript{1,3}\thanks{This work was conducted at Coxwave.}},
    \textbf{Sangyeop Kim\textsuperscript{1,4}\thanks{Corresponding author.}}
    \\
    \small{
        \textsuperscript{1}Coxwave,
        \textsuperscript{2}Kakaobank,
        \textsuperscript{3}Kyung Hee University,
        \textsuperscript{4}Seoul National University
    }
    \\
    \small{
        \texttt{
        yann.lee@kakaobank.com, syw5141@khu.ac.kr, sy917kim@bdai.snu.ac.kr
        }
    }
}

\begin{document}
\maketitle
\begin{abstract}
We present the Conversational Data Retrieval (CDR) benchmark, the first comprehensive test set for evaluating systems that retrieve conversation data for product insights. With 1.6k queries across five analytical tasks and 9.1k conversations, our benchmark provides a reliable standard for measuring conversational data retrieval performance. Our evaluation of 16 popular embedding models shows that even the best models reach only around NDCG@10 of 0.51, revealing a substantial gap between document and conversational data retrieval capabilities. Our work identifies unique challenges in conversational data retrieval (implicit state recognition, turn dynamics, contextual references) while providing practical query templates and detailed error analysis across different task categories. The benchmark dataset and code are available at \url{https://github.com/l-yohai/CDR-Benchmark}.
\end{abstract}

\section{Introduction}
The widespread adoption of generative AI powered by Large Language Models (LLMs) has created vast repositories of conversation data \cite{liang2025widespreadadoptionlargelanguage}. These dialogues offer valuable insights into user behaviors and system performance. However, effectively analyzing and leveraging this accumulated conversational data remains an underexplored challenge in the field \cite{zhang2024imperativeconversationanalysisera, suri2024the, eloundou2025firstperson}.

Unlike conventional information systems, large language model-based systems operate through open-ended interactions without predefined specifications \cite{10.1609/aaai.v38i17.29923, liu2024convbench}. Users interact with them in diverse, unpredictable ways, creating unique challenges for conversational data analytics. Traditional approaches to extracting product insights struggle with these datasets: supervised learning techniques face prohibitive labeling costs \cite{jafarlou2024reducinglabelingcostssentiment}, manual dialogue review becomes impractical at the scale of millions of conversations \cite{bavaresco2020conversational, folstad2021investigating}, and conventional metrics fail to capture the complex evolution of user satisfaction across multiple turns \cite{10.1145/2806416.2806621}.

Given these challenges, many product development teams have adopted an exploratory ``Retrieve and Analyze'' workflow to derive insights from their conversation data (see Appendix \ref{app:business_case} for a detailed real-world case study). In this approach, retrieval quality fundamentally determines analysis effectiveness—if relevant conversations cannot be efficiently found, critical insights remain hidden despite analyst expertise. For example, when investigating satisfaction issues, product managers often use basic keyword searches like ``unhappy'' or ``disappointed,'' missing cases where dissatisfaction is expressed implicitly or across multiple turns. This retrieval gap creates significant blind spots in understanding user experiences and severely limits the value extracted from conversation datasets.

\begin{figure}[h]
    \centering
    \includegraphics[width=1.0\linewidth]{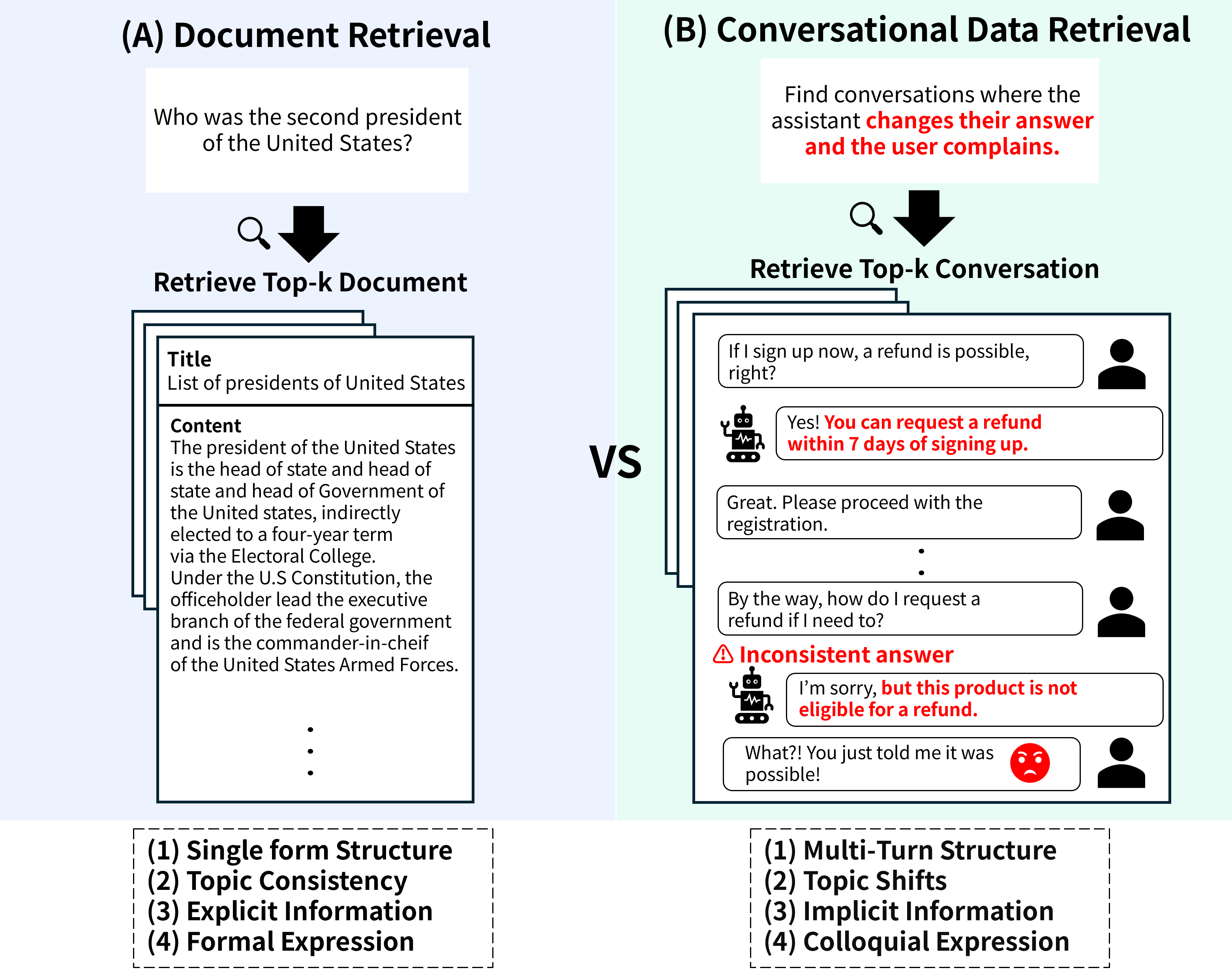}
    \caption{Comparison between traditional document retrieval and conversational data retrieval.}
    \label{fig:key_difference}
\end{figure}

To address these limitations, we build upon the concept of Conversational Data Retrieval (CDR) \cite{HEISIR}: the task of retrieving relevant conversations from large chat histories based on queries targeting conversation-specific content and context. As illustrated in Figure \ref{fig:key_difference}, CDR differs fundamentally from traditional document retrieval \cite{SFRAIResearch2024, zhang2025jasperstelladistillationsota} by addressing conversation-specific challenges: multi-turn exchanges, implicit meanings, and topic shifts.

\begin{figure*}[thbp]
    \centering
    \includegraphics[width=\linewidth]{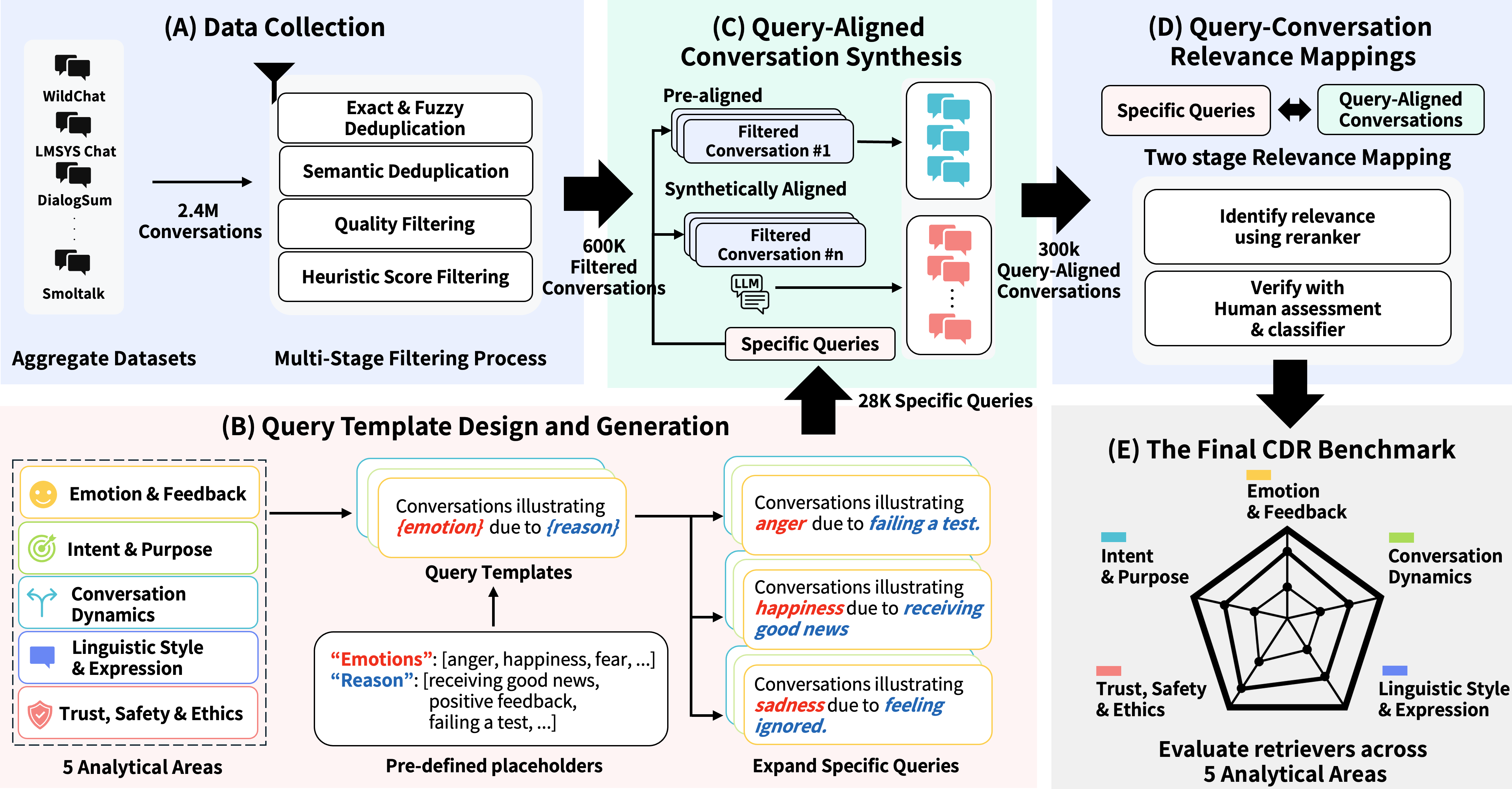}
    \caption{An overview of the Conversational Data Retrieval (CDR) benchmark construction pipeline. (A) Collect and filter large-scale conversational data. (B) Generate query templates across five key areas. (C) Synthesize query-aligned conversations with LLMs. (D) Map relevance through reranking, human assessment, and classifier verification. (E) Integrate the processed data into a standardized CDR evaluation benchmark.}
    \label{fig:main}
\end{figure*}

Beyond analytical use cases, effective CDR enables applications such as AI memory systems \cite{openaimemory, pan2025secom}, and retrieval-augmented generation \cite{10.5555/3495724.3496517, wang2024adaptiveretrievalaugmentedgenerationconversational}. However, current retrieval solutions were not designed with conversations in mind, limiting the potential of these applications.

Despite its value, CDR remains underexplored in research. This gap stems from several factors: the proprietary nature of industrial conversation datasets, privacy concerns limiting public data availability, and the lack of standardized evaluation metrics \cite{10.1145/3209978.3210124, 10.1145/3397271.3401206}. These challenges have hindered research progress on methods specifically designed for CDR.

To address this gap, we introduce a comprehensive benchmark for CDR. Figure \ref{fig:main} illustrates our construction process, including data collection, query design, and validation. Our contributions include: (1) the first benchmark specifically targeting CDR, comprising 1.6k queries and 9.1k conversations; (2) evaluation of 16 commonly used embedding models revealing performance disparities between document and conversation retrieval; (3) a taxonomy of five essential analytical tasks exposing unique challenges in CDR; and (4) practical query templates developed with domain experts applicable to product improvement workflows. This benchmark facilitates structured development of conversation retrieval models, supporting applications and research in conversation analysis.

\section{Related Works}
Information retrieval (IR) has evolved from lexical matching \cite{Robertson2009ThePR} to neural approaches \cite{karpukhin-etal-2020-dense}. Recent advances in generative AI have linked retrieval with conversation, enabling conversational search \cite{Dalton2020CAsT2T, Mao2023LargeLM}, agent memory systems \cite{Packer2023MemGPTTL, Zhong_Guo_Gao_Ye_Wang_2024}, and retrieval-based reasoning \cite{10.1145/3627673.3680087, openaideepresearch}. However, these efforts focus on using retrieval to enhance conversations \cite{Qu2020OpenRetrievalCQ, 10.1145/3580305.3599411}, not on effectively retrieving conversational data itself.

The unique value of conversational data lies in its multifaceted nature. Human-Computer Interaction research has identified several dimensions critical for understanding these interactions—including user intentions, emotional responses, conversation flow patterns, and trust development \cite{bavaresco2020conversational, folstad2021investigating}. These elements often span multiple turns and contain implicit signals that traditional document retrieval approaches struggle to capture \cite{wu2025longmemeval, xue2025mmrc}. Effective analysis requires methods to identify these complex patterns within conversations.

While industrial applications generate vast conversational data, privacy concerns and proprietary issues severely limit public access to these datasets \cite{gumusel2025literature}. Even available datasets often lack sufficient coverage of specific analytical dimensions needed for comprehensive evaluation \cite{zhao2024wildchat, allal2025smollm2smolgoesbig}. 

Synthetic conversational data offers a valuable solution to these constraints, as high-quality synthetic dialogues can match or exceed the performance of systems trained on real data \cite{dai2022dialog, 10.1145/3477495.3531863, wu-etal-2024-synthesizing}. This approach enables more controlled evaluation by systematically varying conversation attributes while maintaining natural dialogue properties.

However, existing IR benchmarks \cite{thakur2021beir, muennighoff-etal-2023-mteb} focus on documents, while dialogue datasets typically focus on generation tasks. This creates a significant gap between the analytical needs identified in HCI research and available evaluation frameworks. Our CDR benchmark addresses this gap by integrating multi-dimensional aspects of conversations with a comprehensive evaluation framework.

\begin{table*}[thbp]
    \centering
    \resizebox{\textwidth}{!}{
    \begin{tabular}{p{5cm}|p{7cm}|p{7cm}}
    \toprule
    \textbf{Analytical Area} & \textbf{Description} & \textbf{Product Insights} \\
    \midrule
    Emotion \& Feedback & Identifying users' emotional states and feedback in conversations & Revealing satisfaction patterns and pain points for product improvement \\
    \midrule
    Intent \& Purpose & Recognizing user intentions and goals & Evaluating alignment between intended and actual AI system usage \\
    \midrule
    Conversation Dynamics & Analyzing conversation flow, turn structure and resolution patterns & Identifying conversation bottlenecks and improving dialogue completion rates \\

    \midrule
    Trust, Safety \& Ethics & Exploring trust-building and ethical issues in conversations & Identifying system reliability concerns and potential safety risks \\
    \midrule
    Linguistic Style \& Expression & Analyzing language patterns and comprehension challenges & Helping calibrate system language to user comprehension levels \\
    \bottomrule
    \end{tabular}
    }
    \caption{Five core analytical areas identified for the CDR benchmark with their product insights.}
    \label{table:analytical-areas}
\end{table*}

\section{Designing the CDR Benchmark}
\paragraph{Data Collection and Industrial Requirements}
To establish a foundation for the CDR benchmark, we collected conversational data from 11 diverse open-source dialogue datasets including LMSYS Chat \cite{10.5555/3692070.3692401}, WildChat \cite{zhao2024wildchat}, and DialogSum \cite{chen-etal-2021-dialogsum}. To ensure quality and remove duplicates, we applied filtering using the NeMo Curator framework \cite{Jennings_NeMo-Curator_a_toolkit}, refining approximately 2.4 million conversations to 600k high-quality dialogue instances. The complete data sources and filtering method are detailed in Appendix \ref{app:data_collection}.

To ensure industrial applicability, we sampled 1k conversations for analysis and gathered input from 20 experts in generative AI product development. From this combined research, we identified key information needs when examining conversational data and determined five core areas for product improvement, shown in Table \ref{table:analytical-areas}. These areas reflect how conversational data differs from traditional document retrieval challenges.

\paragraph{Query Template Design and Generation}
From the five core areas in Table~\ref{table:analytical-areas}, we created 130 query templates that capture the specific characteristics of each category. Each template included placeholder elements to cover diverse conversational scenarios.

For example, a template in the Emotion \& Feedback category might be:
``Find conversations where users express \{emotion\} after \{system\_action\}''
For the \{emotion\} placeholder, values included ``frustration,'' ``disappointment,'' and ``satisfaction.''

We defined approximately 510 placeholder values across different categories. By combining these placeholders with our templates, we generated a total of 28k specific queries. Full details of templates and placeholders are provided in Appendix~\ref{app:query_templates}.

\paragraph{Query-Aligned Conversation Synthesis Method}
Finding conversations that match our diverse queries presented two challenges: our corpus could not cover all specific query scenarios needed, and manually labeling thousands of conversations would be prohibitively time-consuming.

To address these limitations, we first retrieved top-5 candidate conversations for each query using the embedding model \cite{openaiembedding}. Twenty expert annotators with industry experience in conversational AI product development then manually reviewed these candidates. They classified each as related or unrelated based on whether it faithfully reflected the query intent. When no suitable match existed, we used reasoning-capable language models—Claude-3.7 \cite{claude37sonnet}, o1 \cite{openaio1}, and o3-mini \cite{openaio3mini}—to create synthetically aligned conversations by adapting existing conversations from our corpus. These LLM-generated conversations were also validated by expert annotators to ensure both query fidelity and conversational naturalness.

Our conversation generation prompt (detailed in Appendix \ref{app:conv-gen-prompt}) instructed models to maintain each conversation structure and characteristics while incorporating elements needed for query alignment. This approach preserved the natural variation found in real conversations while ensuring examples contained features necessary for evaluation.

\begin{figure}[htbp]
    \centering
    \includegraphics[width=\linewidth]{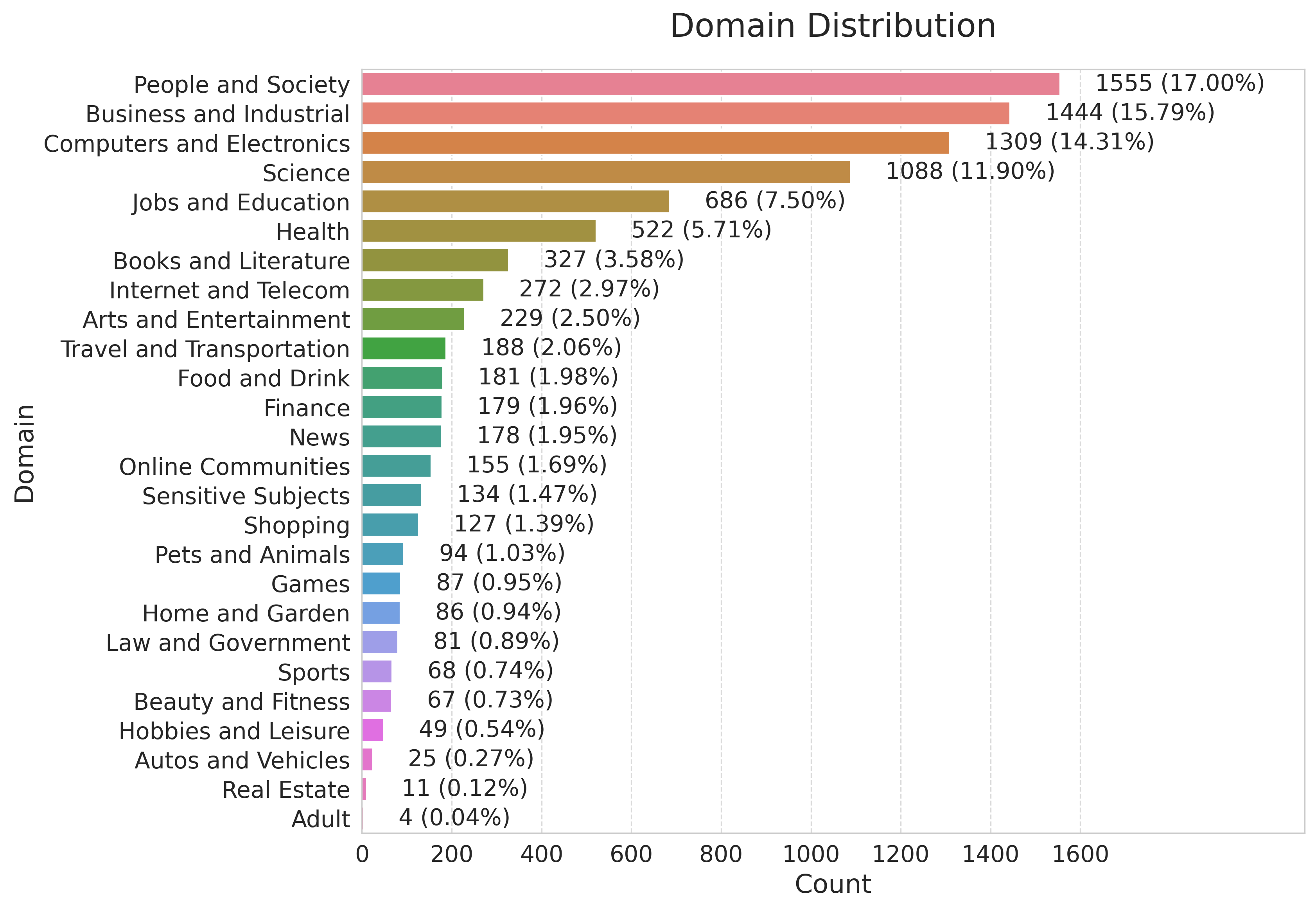}
    \caption{Domain distribution in the CDR benchmark dataset, showing diverse coverage across categories.}
    \label{fig:domain-distribution}
\end{figure}

By combining pre-aligned conversations with synthetically aligned conversations, our method maintained domain diversity. Figure \ref{fig:domain-distribution} shows balanced coverage across major categories like People and Society (17.00\%) and Business and Industrial (15.79\%), as classified by a fine-tuned 
classifier\footnote{\url{https://huggingface.co/nvidia/domain-classifier}}.

\paragraph{Expanding Query-Conversation Relevance Mappings}

To create a realistic retrieval benchmark, we expanded each query to match multiple relevant conversations through a three-step process. First, we trained a specialized reranker model using 300k conversations from our corpus. We generated positive and negative query examples with LLaMa 3.3 70B \cite{llama3370b} using prompts in Appendix~\ref{app:query-generation-prompt} and \ref{app:query-aug-prompt}, and fine-tuned the GTE Reranker \cite{zhang-etal-2024-mgte} (training detailed in Appendix \ref{app:reranker_training}). Second, we applied this reranker to identify candidate relevant conversations, selecting pairs with relevance scores above 0.9 and excluding overly general queries matching more than 50 conversations.

Third, we validated mappings through comprehensive human assessment. Expert annotators conducted full manual evaluation of approximately 4k query-conversation pairs across 200 queries. We applied binary relevance criteria with majority voting for reliability, conservatively removing non-consensus cases. We trained a ModernBERT-based \cite{warner2024smarterbetterfasterlonger} relevance classifier using these manually validated pairs, achieving 95.2\% accuracy, as detailed in Appendix \ref{app:classifier_training}. For remaining queries, we applied the classifier to predict relevance for all pairs, then employed two-stage human verification. First, we prioritized uncertain cases where sigmoid scores fell below 0.9. Second, we identified boundary inconsistencies where irrelevant predictions appeared among relevant pairs, and vice versa. Human annotators verified both uncertain predictions and inconsistent boundaries, ensuring comprehensive coverage while efficiently allocating annotation effort to critical cases.

\begin{table}[htbp]
\centering
\resizebox{0.65\columnwidth}{!}{%
\begin{tabular}{l|r}
\toprule
\multicolumn{2}{c}{\textbf{General Statistics}} \\
\midrule
Number of conversations & 9,146 \\
Number of queries & 1,583 \\
Avg. messages per conversation & 5.4 \\
Avg. tokens per conversation & 464 \\
Avg. relevant convs per query & 20.44 \\
Total query-conversation pairs & 32,357 \\
\midrule
\multicolumn{2}{c}{\textbf{Query Task Distribution (\%)}} \\
\midrule
Intent \& Purpose & 36.1\% \\
Emotion \& Feedback & 20.1\% \\
Linguistic Style \& Expression & 15.9\% \\
Trust, Safety \& Ethics & 14.6\% \\
Conversation Dynamics & 13.4\% \\
\bottomrule
\end{tabular}%
}
\caption{Key statistics of the CDR benchmark dataset.}
\label{table:cdr-statistics}
\end{table}

\paragraph{Benchmark Composition and Characteristics}

Our comprehensive mapping pipeline provides an efficient method for constructing high-quality query-conversation pairs. This methodology offers a practical solution for industrial deployment where cost-effective data mapping is essential. However, as our goal is to establish a rigorous benchmark, we conducted additional validation to ensure maximum integrity. We employed four LLMs—GPT-4o \cite{openai2024gpt4ocard}, o3-mini \cite{openaio3mini}, Claude 3.7 Sonnet \cite{claude37sonnet}, and Gemini 2.0 Pro \cite{gemini20pro}—with the prompt in Appendix~\ref{app:rel-clf-prompt} to cross-check all pairs. Cases where LLMs disagreed were flagged for expert review by annotators, who applied consistent binary relevance criteria with majority voting. Pairs without clear consensus were conservatively discarded. Through this multi-stage validation approach combining LLM scalability with human verification at each step, 97\% of all query-conversation mappings passed assessment, ensuring the final benchmark meets the highest quality standards.

The final CDR benchmark consists of 1,583 queries and 9,146 conversations (Table~\ref{table:cdr-statistics}). Conversations average 5.4 messages and 464 tokens\footnote{Based on the GPT-4o tokenizer.}. Each query maps to 20.44 relevant conversations on average. Query distribution spans five core areas: Intent \& Purpose (36.1\%), Emotion \& Feedback (20.1\%), and three other categories, with detailed examples provided in Appendix~\ref{app:query_conversation_examples}.

\begin{table*}[thbp]
    \centering
    \begin{adjustbox}{max width=\textwidth}
    \renewcommand{\arraystretch}{1.1}
    \begin{tabular}{l c c c c c c c c c c c}
        \toprule[1.2pt]
        \multirow{2}{*}{\textbf{Model}} & \multicolumn{3}{c}{\textbf{Turn}} & \multicolumn{3}{c}{\textbf{Sliding chunk (k=3)}} & \multicolumn{3}{c}{\textbf{Session}} & \multirow{2}{*}{\textbf{Inference (s)}} & \multirow{2}{*}{\textbf{Ingestion (s)}} \\
        \cmidrule(lr){2-4} \cmidrule(lr){5-7} \cmidrule(lr){8-10}
        \multicolumn{1}{c}{} & \textbf{NDCG@10} & \textbf{R@10} & \textbf{P@10} & \textbf{NDCG@10} & \textbf{R@10} & \textbf{P@10} & \textbf{NDCG@10} & \textbf{R@10} & \textbf{P@10} & & \\
        \midrule[1pt]
        \rowcolor{gray!15} \multicolumn{12}{l}{\textbf{Commercial API Models}} \\
        \cmidrule[0.8pt]{1-12}
        Voyage-3-large & \textbf{0.5079} & \underline{0.2609} & \underline{0.4359} & \underline{0.5063} & \underline{0.2582} & \underline{0.4327} & \textbf{0.5036} & \textbf{0.2615} & \textbf{0.4358} & 375.03 & 2620.00 \\
        \midrule[0.3pt]
        Text-embedding-3-large & \underline{0.5078} & \textbf{0.2698} & \textbf{0.4389} & \textbf{0.5130} & \textbf{0.2696} & \textbf{0.4416} & \underline{0.4876} & \underline{0.2529} & \underline{0.4190} & 245.90 & 1433.12 \\
        Text-embedding-3-small & 0.4897 & 0.2558 & 0.4183 & 0.4855 & 0.2558 & 0.4171 & 0.4664 & 0.2412 & 0.3972 & \underline{196.99} & \underline{1253.68} \\
        \midrule[0.3pt]
        Embed-english-v3.0 & 0.4189 & 0.2237 & 0.3547 & 0.2547 & 0.1351 & 0.2116 & 0.3620 & 0.1923 & 0.2987 & \textbf{83.75} & \textbf{520.00} \\
        \midrule[1pt]
        \rowcolor{gray!15} \multicolumn{12}{l}{\textbf{Open Source Models}} \\
        \cmidrule[0.8pt]{1-12}
        Stella\_en\_1.5B\_v5 & \textbf{0.4907} & \textbf{0.2592} & \textbf{0.4141} & \textbf{0.4894} & \textbf{0.2528} & \textbf{0.4078} & \textbf{0.4722} & \textbf{0.2481} & \textbf{0.3961} & 7.84 & 336.92 \\
        Stella\_en\_400M\_v5 & \underline{0.4682} & \underline{0.2490} & \underline{0.3963} & \underline{0.4651} & \underline{0.2462} & \underline{0.3919} & 0.4583 & \underline{0.2400} & 0.3846 & 5.51 & 119.24 \\
        \midrule[0.3pt]
        Jasper\_en\_vision\_language\_v1 & 0.4379 & 0.2317 & 0.3712 & 0.4309 & 0.2245 & 0.3615 & 0.4561 & 0.2382 & 0.3814 & 7.86 & 355.56 \\
        \midrule[0.3pt]
        NV-Embed-v2 & 0.3170 & 0.2008 & 0.3251 & 0.3956 & 0.1988 & 0.3262 & \underline{0.4592} & 0.2344 & \underline{0.3855} & 13.92 & 279.73 \\
        NV-Embed-v1 & 0.2467 & 0.1226 & 0.1956 & 0.2603 & 0.1302 & 0.2080 & 0.4389 & 0.2242 & 0.3634 & 13.79 & 280.23 \\
        \midrule[0.3pt]
        SFR-Embedding-2\_R & 0.3344 & 0.1775 & 0.2805 & 0.3127 & 0.1639 & 0.2589 & 0.4474 & 0.2280 & 0.3722 & 10.40 & 213.57 \\
        \midrule[0.3pt]
        Jina-embeddings-v3 & 0.3803 & 0.2053 & 0.3160 & 0.3983 & 0.2142 & 0.3363 & 0.3718 & 0.1995 & 0.3088 & 7.01 & 106.25 \\
        \midrule[0.3pt]
        Modernbert-embed-base & 0.3594 & 0.1923 & 0.3026 & 0.3398 & 0.1795 & 0.2857 & 0.3579 & 0.1906 & 0.3016 & 7.49 & \underline{45.63} \\
        \midrule[0.3pt]
        Gte-Qwen2-1.5B-instruct & 0.4646 & 0.2412 & 0.3952 & 0.4386 & 0.2261 & 0.3708 & 0.3615 & 0.1919 & 0.2987 & 7.59 & 336.73 \\
        Gte-large-en-v1.5 & 0.3310 & 0.1821 & 0.2792 & 0.3246 & 0.1778 & 0.2726 & 0.3429 & 0.1840 & 0.2860 & \underline{4.39} & 133.29 \\
        \midrule[0.3pt]
        Bge-large-en-v1.5 & 0.3276 & 0.1757 & 0.2719 & 0.3105 & 0.1659 & 0.2539 & 0.3071 & 0.1617 & 0.2476 & \textbf{3.97} & 94.29 \\
        \midrule[0.3pt]
        Cde-small-v2 & 0.1163 & 0.0606 & 0.0975 & 0.1226 & 0.0640 & 0.1007 & 0.0830 & 0.0463 & 0.0701 & 7.50 & \textbf{39.57} \\
        \bottomrule[1.2pt]
    \end{tabular}
    \end{adjustbox}
    \caption{Performance comparison of 16 embedding models on the CDR benchmark. The table presents retrieval metrics including NDCG@10, Recall@10 (R@10), and Precision@10 (P@10) across three evaluation settings: Turn-based, Sliding chunk (k=3), and Session-based approaches. Additional metrics are in Appendix~\ref{apdx:additional_metric}. Inference and ingestion times were measured using procedures in Appendix~\ref{app:evaluation_efficiency} on hardware described in Appendix~\ref{app:evaluation_spec}.}
    \label{table:embedding-model-result}
\end{table*}

\section{Experiments and Analysis}

\subsection{Experimental Setup}
We evaluated 16 widely used embedding models from open-source communities and commercial providers including OpenAI \cite{openaiembedding}, Cohere \cite{embedv3}, and Voyage AI \cite{voyage3large}. Performance was assessed using NDCG@10, Recall@10, and Precision@10 at three retrieval granularities—session-level, turn-level, and sliding window (chunk size = 3). For detailed evaluation methodology, see Appendix~\ref{app:evaluation_setup}.

\subsection{Results and Analysis}

\begin{table*}[thbp]
    \centering
    \renewcommand{\arraystretch}{0.8}
    \fontsize{7.4pt}{10pt}\selectfont
    \begin{tabular}{p{2.3cm}|p{2.3cm}|p{6.1cm}|p{3.3cm}}
        \toprule
        \textbf{Challenge Type} & \textbf{Query Example} & \textbf{Incorrectly Retrieved Results} & \textbf{Why Models Fail} \\
        \midrule
        Role Recognition \newline Failure & Assistant shares parenting and childcare advice &
        \textbf{user}: Welcome to the parent teacher conference. So what is your child's name?\newline
        \textbf{assistant}: Megan Jones.\newline
        \textbf{user}: She's been having some problems with the other kids in your class.
        & Models match ``parent,'' ``child,'' ``teacher'' keywords but miss conversational roles. Assistant is receiving information as parent, not providing advice. \\ \midrule
        Dynamic Progression Failure & Conversation where user feels increasingly satisfied with assistant & \textbf{user}: You've been so helpful with all my questions lately. I just wanted to tell you how happy I am with your assistance.\newline
        \textbf{assistant}: Thank you so much for your kind words! It truly means a lot to me.
        & Models match final satisfaction but miss progressive ``increasingly'' aspect. This shows static state, not gradual improvement. \\ \midrule
        Semantic Contextual Misinterpretation & Assistant provides real estate and housing information & \textbf{user}: I'm visiting friends in Nairobi. What's the weather like?\newline
        \textbf{assistant}: 103°F, 2\% chance of rain.\newline
        \textbf{user}: I need a house for 1 with laundry service.\newline
        \textbf{assistant}: Found a house at Chiromo Road with 4.6 rating.
        & Models match "house" keyword but miss context. This is travel booking service, not real estate information provision. \\ 
        \bottomrule
    \end{tabular}
    \caption{Representative failure cases illustrating major retrieval challenges in conversation understanding tasks.}
    \label{table:failure_cases}
\end{table*}

Table \ref{table:embedding-model-result} summarizes model performance on our CDR benchmark. Among commercial API models, Voyage-3-large \cite{voyage3large} achieved the highest performance in both turn-based (NDCG@10: 0.5079) and session-based (NDCG@10: 0.5036) evaluation, while Text-embedding-3-large \cite{openaiembedding} led in sliding chunk settings (NDCG@10: 0.5130).
Among open-source models, Stella\_en\_1.5B\_v5 \cite{zhang2025jasperstelladistillationsota} demonstrated consistently high performance across all evaluation settings.
Interestingly, some models showed significant performance variations across different evaluation settings. For instance, NV-Embed-v2 performed poorly in turn-based evaluation (NDCG@10: 0.3170) but achieved substantially improved performance in session-based evaluation (NDCG@10: 0.4592).
Even top-performing models scored just above 0.5 in NDCG@10, highlighting the challenges of modeling conversational structure, context transitions, and implicit references.

\begin{figure}[!htbp]
    \centering
    \includegraphics[width=0.98\linewidth]{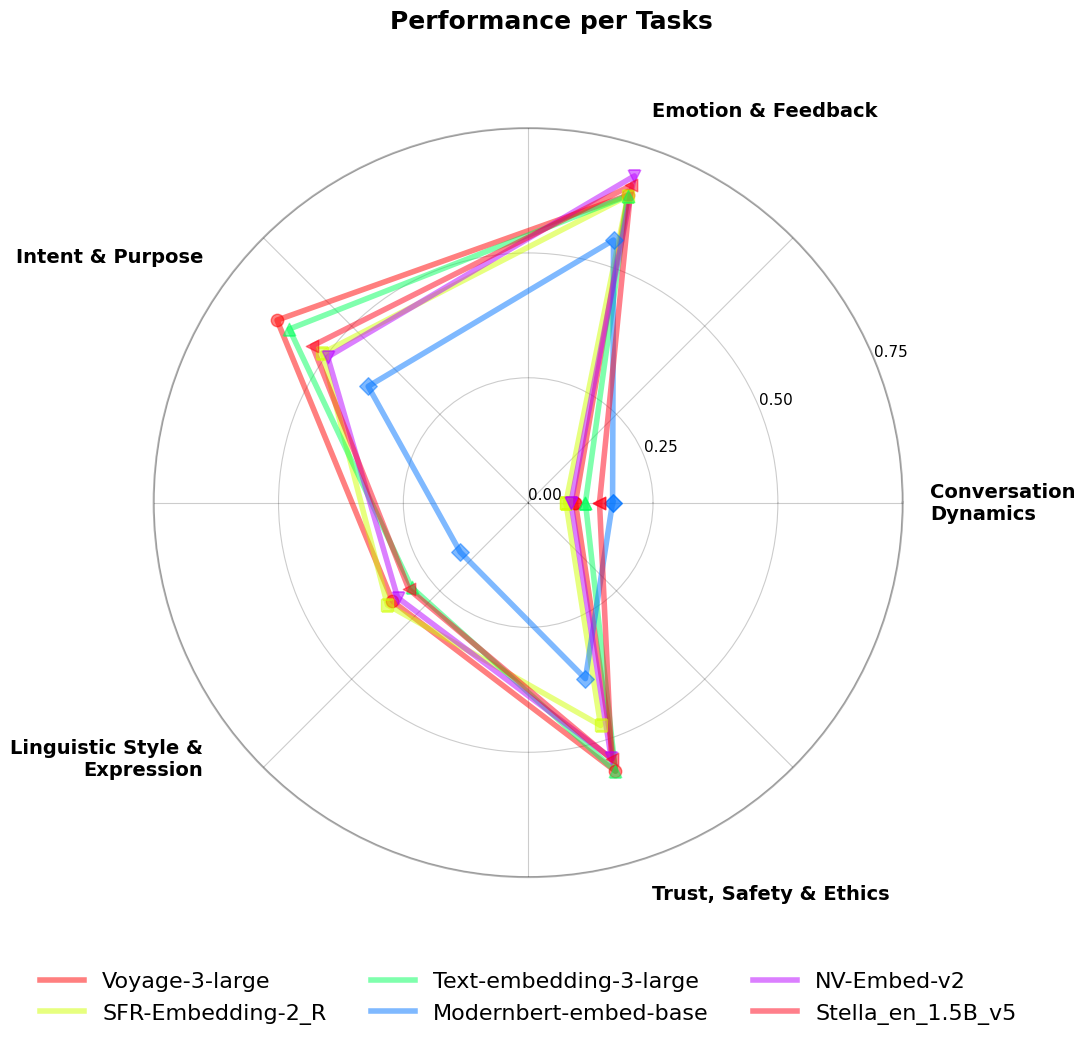}
    \caption{Task-specific NDCG@10 performance comparison of top-performing embedding models and category winners. All results are available in Appendix \ref{apdx:all_result}.}
    \label{fig:result-per-task}
\end{figure}

\subsection{Performance Across Task Categories}

Figure \ref{fig:result-per-task} reveals performance variations across task types. All models score highest in `Emotion \& Feedback' and `Intent \& Purpose', but perform poorly in `Conversation Dynamics' where even the best models score below 0.17. This suggests current models are good at understanding content and explicit statements but struggle with understanding how conversations develop and flow. 

No model excels across all categories - even top-performing Voyage-3-large shows varied results. This suggests no dominant approach exists yet for CDR. Optimal architectures remain unexplored, particularly for conversation structure understanding, which is crucial for practical applications.

\subsection{Analysis of Retrieval Failures}

We identified three consistent failure patterns in current embedding approaches to CDR. Table~\ref{table:failure_cases} illustrates these challenges: \textit{Role Recognition Failure}, \textit{Dynamic Progression Failure}, and \textit{Semantic Contextual Misinterpretation}. These failures occur across both turn-based and session-based analysis, revealing models' inability to capture conversation dynamics at multiple levels. The consistent challenge is that models miss implicit meanings that emerge from conversational context—patterns where the actual roles, progressive changes, or situational context must be inferred from dialogue flow rather than explicit keyword matching.

These errors stem from a fundamental limitation: current models process conversations as collections of words and topics similar to documents, rather than as dynamic exchanges with temporal flow and implicit state changes. Standard embeddings capture vocabulary similarities but miss the contextual evolution and interactive nature of dialogue. This explains the poor performance in \textit{Conversation Dynamics} across all models and signals that effective retrieval systems must be redesigned to capture the unique properties of human dialogue like turn-taking patterns and implicit state transitions.

\section{Conclusion}

Conversational Data Retrieval (CDR) benchmark establishes the first comprehensive framework for evaluating retrieval systems on conversation data. Experimental results show that even the highest-performing models have not reached satisfactory performance. Our benchmark exposes fundamental challenges unique to conversational data: understanding implicit states, tracking conversation flow, and interpreting contextual references. Our work provides standardized evaluation methodology and query templates for product improvement while establishing a foundation for conversation-specific retrieval techniques that better capture the multi-dimensional nature of human-AI interactions.

\section*{Limitations}
Our benchmark is limited to English text-based conversations, which may constrain evaluation in multilingual or multimodal settings. This focus, while enabling controlled evaluation, could limit the broader applicability of our findings to diverse linguistic contexts and interaction modalities in global conversational AI applications.

Our benchmark evaluates embedding-based retrieval models, reflecting their widespread adoption in conversational memory systems where turn, session, and segment-level granularities are commonly employed. However, the lack of specialized retrieval models designed specifically for conversation represents a gap in the field that our benchmark could help address through future development of conversation-tailored representation architectures.

While our benchmark provides comprehensive evaluation of retrieval models with robust data quality validation through domain expert involvement, it does not extend to empirical studies of industrial problem-solving applications. Although our motivation stems from real-world challenges and our benchmark identifies optimal approaches under current conditions, further research is needed to validate the practical value of these findings in actual deployment scenarios and their impact on end-user satisfaction in conversational AI systems.

\section*{Ethical Considerations}
In the development and application of the CDR benchmark, we carefully considered various ethical aspects. Since conversational data inherently contains user interactions and diverse linguistic expressions, we prioritized privacy protection throughout the data collection and processing stages. We utilized only publicly available open-source datasets and included specific guidelines in our conversation generation prompts to address any potentially remaining personal expressions or sensitive information (see Appendix \ref{app:license}). These guidelines included instructions to ``appropriately redact or anonymize any PII in reference conversations,'' ``avoid generating conversations that could be misleading, harmful, or promote unethical behavior,'' and ``ensure that no personally identifiable information such as names, addresses, phone numbers, financial details, social security numbers, or other sensitive data is exposed or inferred.''

To ensure diversity and balance in the conversational data, we designed the benchmark dataset to encompass a wide range of domains without bias toward specific topics or areas. As shown in Figure \ref{fig:domain-distribution}, we included a balanced representation of conversations from broad domains such as society, business, technology, and science, thereby minimizing bias toward particular areas. Additionally, we explicitly incorporated a `Trust, Safety \& Ethics' category in the benchmark's task areas to establish ethical conversational retrieval capabilities as an important evaluation criterion.

While conversational data retrieval technology can contribute to positive purposes such as improving service quality and user experience, it also carries potential risks of privacy infringement or misuse as inappropriate surveillance tools. We recognize this duality and hope that the CDR benchmark will serve as a tool to promote balance between ethical values and innovative technological advancement. Through this, we believe that the development of conversational AI systems can progress in a direction that respects users' rights and dignity.

\section*{Acknowledgments}

This work was supported by Coxwave, Artificial intelligence industrial convergence cluster development project funded by the Ministry of Science and ICT(MSIT, Korea) \& Gwangju Metropolitan City, Institute for Information \& communications Technology Planning \& Evaluation(IITP) grant funded by the Korea government(MSIT) (RS-2019-II190075, Artificial Intelligence Graduate School Program(KAIST)) and the Korea government(MSIT) (No. RS-2024-00509279, Global AI Frontier Lab).

\bibliography{custom}
\newpage
\appendix
\section{Industry Case Study of the ``Retrieve and Analyze'' Approach}
\label{app:business_case}
The ``Retrieve and Analyze'' methodology mirrors how analysts and product managers naturally approach problem-solving in real business environments. When faced with user feedback or product issues, human analysts typically form initial hypotheses, gather relevant examples, analyze patterns, and progressively refine their understanding through iterative investigation. What has changed with large-scale conversation data is not this fundamental analytical process, but rather the need for computational assistance to efficiently navigate thousands or millions of conversations.

The following case study illustrates how this human-centered analytical approach, supported by conversation retrieval capabilities, works in practice. This example is adapted from an actual business scenario at a health and fitness application company, demonstrating both the natural analytical workflow and the positive impact of effective conversational data retrieval.

A customer experience team was investigating increased user dissatisfaction following a recent update. Traditionally, they relied on user ratings and manual reviews of customer complaints. However, after implementing chatbot support, these methods became inadequate - the chatbot interface lacked rating systems, and the chat volume overwhelmed manual inspection capabilities.

To address this challenge, the team implemented a conversation retrieval system, beginning with a broad query: ``Find sessions where users express dissatisfaction.'' Sample analysis revealed mentions of the points reward system, prompting them to deep dive with a more targeted search: ``Find conversations where users express dissatisfaction with changes to the points reward system'' to determine if this was a widespread issue rather than isolated incidents.

This refined approach confirmed their hypothesis, revealing reduced point accumulation rates as the primary driver of dissatisfaction, with users consistently comparing the new system unfavorably to the previous one. Through this methodical process of hypothesis formation and targeted validation, the team efficiently pinpointed the specific issue causing user frustration—a discovery that would have consumed significantly more time and resources using traditional review methods.

\section{Dataset Sources and Filtering Method}
\label{app:data_collection}

\subsection{Dataset Sources}

We constructed an initial dataset comprising around 2.4 million conversations by aggregating 11 diverse open-source datasets. To ensure broad coverage of dialogue scenarios, our dataset includes both real-world and synthetic conversational data. An overview of the datasets is provided in Table~\ref{tab:conversation_datasets}.

\begin{table*}[htbp]
    \centering
    \small
    \begin{tabular}{m{4.2cm}|m{1.8cm}|m{7cm}}
        \toprule
        \textbf{Dataset} & \textbf{Data Size} & \textbf{Key Features} \\
        \midrule
        \rowcolor{gray!15} \multicolumn{3}{l}{\textbf{Real-world Conversation Data}} \\
        \midrule
        LMSYS-Chat-1M \newline \cite{zheng2024lmsyschatm} & 1,000,000+ & 
        Real-world user-LLM chats; multi-turn; multilingual; moderation tags and PII redacted \\
        \midrule
        WildChat-1M \newline \cite{zhao2024wildchat} & 1,000,000+ & 
        User-ChatGPT logs; multilingual; includes user metadata and toxicity labels \\
        \midrule
        DialogSum \newline  \cite{chen-etal-2021-dialogsum} & 12,000+ & 
        Real-world conversations; paired with abstractive summaries and topic annotations \\
        \midrule
        DailyDialog \newline  \cite{li-etal-2017-dailydialog} & 10,000+ & 
        Open-domain daily conversations; annotated with dialogue acts and emotions \\
        \midrule
        MultiWOZ 2.2 \newline  \cite{zang-etal-2020-multiwoz} & 8,000+ & 
        Multi-domain, task-oriented dialogues; annotated with states and system actions; corrected labels \\
        \midrule
        Bot-Adversarial Dialogue (BAD) \newline  \cite{xu-etal-2021-bot} & 5,000+ & 
        Adversarial conversations to test chatbot safety; includes persona settings and safety labels \\
        \midrule
        MobileConvRec \newline \cite{maji2024mobileconvrecconversationaldatasetmobile} & 8,000+ & 
        Conversations for mobile app recommendation; multi-turn; includes user feedback and app info \\
        \midrule
        OpenDialKG \newline  \cite{moon-etal-2019-opendialkg} & 12,000+ & 
        Knowledge-grounded conversations; each turn linked to KG entities for explainability \\
        \midrule
        \rowcolor{gray!15} \multicolumn{3}{l}{\textbf{Synthetic Conversational Data}} \\
        \midrule
        SmolTalk \newline \cite{allal2025smollm2smolgoesbig} & 1,000,000+ & 
        Synthetic dialogues for instruction following; wide coverage (QA, summarization, coding tasks) \\
        \midrule
        Bitext Customer Support \newline \cite{bitext_customer_support_dataset} & 26,000+ & 
        Synthetic QA pairs created by linguists; customer support domain; slot annotations \\
        \midrule
        Schema-Guided Dialogue (SGD) \newline  \cite{rastogi2020towards} & 16,000+ & 
        Multi-domain task-oriented dialogues; annotated with intents, slots, states; includes zero-shot domains \\
        \bottomrule
    \end{tabular}
    \caption{Summary of public conversational datasets.}
    \label{tab:conversation_datasets}
\end{table*}

\subsection{Filtering Method}
\label{app:filtering}
For data quality management, we employed a multi-stage filtering process using the NeMo Curator framework. We first applied exact and fuzzy deduplication to remove identical or near-identical conversations. Next, we conducted semantic deduplication by utilizing a model fine-tuned for semantic search\footnote{\url{https://huggingface.co/sentence-transformers/all-MiniLM-L6-v2}}, effectively filtering out semantically redundant instances. For quality filtering, we employed a model fine-tuned for conversation quality assessment\footnote{\url{https://huggingface.co/nvidia/quality-classifier-deberta}}, retaining only conversations labeled as high-quality among high, middle, and low categories. Finally, heuristic score filtering was applied to remove data with excessive punctuation, URLs, and repeated lines, paragraphs, or $n$-grams. Through this comprehensive filtering process, we obtained approximately 600,000 refined conversation data points from an initial set of 2.4 million dialogues.

\section{Template and Placeholders}
\label{app:query_templates}
To ensure the CDR benchmark captures realistic analytical scenarios, we collaborated with domain experts in conversational AI product development to design comprehensive query templates and placeholder values. This expert-guided approach ensures that our benchmark reflects actual information needs encountered in industrial applications. All templates and placeholder values are available in our public repository.

\subsection{Query Template}
We provide representative examples of query templates used in the benchmark generation across five task categories (\textit{Emotion \& Feedback, Intent \& Purpose, Conversation Dynamics, Trust, Safety \& Ethics, Linguistic Style \& Expression}). 
The examples of these query templates are presented in Table~\ref{tab:query-template-examples}.

\begin{table*}[htbp]
\centering
\small
\begin{tabular}{m{0.25\textwidth} | m{0.65\textwidth}}
\toprule
\textbf{Task Category} & \textbf{Query Template Examples} \\ 
\midrule
\textbf{Emotion \& Feedback} & 
\begin{itemize}[noitemsep]
    \item Conversation ending in \{emotion\}.
    \item Conversation reflecting \{emotion\} in \{user\_role\}'s feedback.
    \item Conversation where \{user\_role\} finds \{assistant\_role\}'s answers inadequate, leading to \{emotion\}.
\end{itemize} \\ \hline
\textbf{Intent \& Purpose} & 
\begin{itemize}[noitemsep]
    \item Conversation where \{user\_role\} seeks advice on \{intent\}.
    \item Conversation expressing gratitude to \{assistant\_role\} for resolving \{intent\}.
    \item Conversation where \{user\_role\} repeatedly asks for clarification on \{intent\}.
\end{itemize} \\ \hline
\textbf{Conversation Dynamics} & 
\begin{itemize}[noitemsep]
    \item Conversation where \{user\_role\} becomes more frustrated over time.
    \item Conversation where \{user\_role\} shifts from skepticism to trust.
    \item Conversation where \{user\_role\} feels a sense of accomplishment after resolving an issue with \{assistant\_role\}.
\end{itemize} \\ \hline
\textbf{Trust, Safety \& Ethics} & 
\begin{itemize}[noitemsep]
    \item Conversation expressing distrust to \{assistant\_role\}.
    \item Conversation where \{user\_role\} raises a \{concern\_type\} issue.
    \item Conversation where \{user\_role\} questions the ethics of \{assistant\_role\}.
\end{itemize} \\ \hline
\textbf{Linguistic Style \& Expression} & 
\begin{itemize}[noitemsep]
    \item Conversation using \{linguistic\_style\} in \{user\_role\}'s questions.
    \item Conversation highlighting \{expression\_type\} in \{assistant\_role\}'s feedback.
    \item Conversation where \{assistant\_role\} uses \{linguistic\_style\} to simplify concepts.
\end{itemize} \\
\bottomrule
\end{tabular}
\caption{Examples of Query Templates.}
\label{tab:query-template-examples}
\end{table*}

\vspace{1.5cm}

\subsection{Placeholders and Example Values}
We illustrate examples of placeholder values utilized within query templates in Table~\ref{tab:placeholder-values}, showing potential variability across queries generated for the benchmark.

\begin{table*}[htbp]
\centering
\small
\renewcommand{\arraystretch}{1.2}
\begin{tabular}{m{0.25\textwidth} | m{0.68\textwidth}}
\toprule
\textbf{Placeholder} & \textbf{Values} \\ 
\midrule
\textbf{emotion} & \scriptsize{anger, happiness, fear, sadness, disgust, surprise} \\ \hline
\textbf{reason} & \scriptsize{receiving good news, achieving a goal, success in a project, positive feedback, unexpected reward, losing an opportunity, failing a test, getting rejected, career setback, missed deadline, miscommunication, argument with a friend, relationship conflict, family issues, betrayal, overwhelming workload, financial stress, health concerns, uncertainty about the future, burnout, feeling ignored, being misunderstood, lack of appreciation, being left out, social anxiety, unexpected kindness, support from a friend, acts of generosity, reunion with a loved one, forgiving someone, public embarrassment, making a mistake, feeling inadequate, past regrets, personal failure, exploring a new hobby, intellectual curiosity, inspiring conversation, learning something new, self-discovery, bad weather, technical difficulties, traffic jam, missed appointment, unexpected delay, change in routine, relocation to a new place, adjusting to a new culture, meeting new people, losing a loved one, receiving criticism, feeling judged, comparison with others, unmet expectations, fear of failure, unexpected surprise, random compliment, winning a competition, realizing personal growth, achieving recognition} \\ \hline
\textbf{linguistic\_style} & \scriptsize{formal, informal, neutral, technical, emotional, direct, indirect, logical, persuasive, descriptive, concise, elaborate, colloquial, humorous, sarcastic, empathetic, diplomatic, instructional, academic, poetic, authoritative, friendly, supportive, motivational, analytical, objective, subjective, casual, metaphorical, rhetorical, minimalist, detailed, straightforward, evocative, apologetic, provocative, encouraging, critical, optimistic, pessimistic} \\ \hline
\textbf{expression\_type} & \scriptsize{descriptive, interrogative, exclamatory, imperative, figurative, humorous, sarcastic, rhetorical, analytical, persuasive, ironic, metaphorical, hyperbolic, understated, concise, elaborate, critical, supportive, enthusiastic, skeptical, neutral, emotional, empathetic, diplomatic, apologetic, provocative, assertive, tentative, cautious, objective, subjective, optimistic, pessimistic, directive, expressive, reflective, affirmative, defensive} \\ \hline
\textbf{concern\_type} & \scriptsize{technical issue, ethical issue, academic concern, personal dilemma, relationship issue, work-related stress, health concern, financial problem, social issue, philosophical question, legal complication, moral dilemma, psychological distress, political concern, environmental issue, cultural conflict, safety concern, privacy issue, existential crisis, career uncertainty, education challenge, family dispute, mental health struggle, identity crisis, communication breakdown, trust issue, decision-making difficulty, peer pressure, unfair treatment, discrimination concern, technology misuse, misinformation problem, data security risk, work-life balance struggle, burnout risk, lack of recognition, fear of failure, fear of rejection, self-doubt, unmet expectations, social anxiety, public speaking fear, future uncertainty, innovation challenge, unresolved conflict, resource limitation, competitiveness pressure, time management struggle, productivity concern} \\ \hline
\textbf{information\_type} & \scriptsize{definition, example, guideline, principle, theory, framework, explanation, best practice, case study, historical background, technical specification, algorithm, code snippet, data analysis, statistical insight, latest trend, research finding, scientific evidence, hypothesis, methodology, comparison, contrast, step-by-step guide, practical tip, troubleshooting guide, expert opinion, prediction, future outlook, risk assessment, ethical consideration, common misconception, application, use case, feasibility study, performance evaluation, benchmarking result, legal implication, policy overview, economic impact, market analysis, psychological insight, philosophical perspective, security risk, data privacy issue, innovation strategy, optimization technique} \\ \hline
\textbf{intent} & \scriptsize{Definition Query, Factual Query, How-to Query, Comparison Query, Reason and Consequence Query, Current Events Query, Historical Query, New Service Request, Purchase and Order Placement, Reservation and Booking, Account Creation and Management, Subscription and Membership, Payment Processing, Technical Troubleshooting, Account Recovery and Access Issues, Product Usage Guidance, Service Interruption Support, Complaint Handling, Return and Refund Assistance, Post-Purchase Support, Service Modification, Profile Update, Customization Request, Recommendation Request, Miscellaneous, Greeting, Farewell, Agreement or Acceptance, Disagreement or Rejection, Clarification Request, Repetition Request, Miscellaneous, Content Creation, Content Editing, Brainstorming and Idea Generation, Content Organization, Content Analysis, Miscellaneous, Educational Query, Skill Development, Health and Wellness, Miscellaneous, Positive Emotion Towards Chatbot, Negative Emotion Towards Chatbot, Positive Emotion About Personal Situation, Negative Emotion About Personal Situation, Positive Emotion About External Situation, Negative Emotion About External Situation, Miscellaneous, Offensive Language, Prohibited Content, Malicious Behavior, Miscellaneous} \\ \hline
\textbf{issue\_description} & \scriptsize{technical malfunction, algorithmic bias, ethical dilemma, unexpected software bug, unclear instructions, ambiguous response, miscommunication, incomplete explanation, contradictory information, unresolved question, flawed reasoning, lack of supporting evidence, data inconsistency, security vulnerability, privacy violation, inaccurate prediction, unmet expectations, slow response time, unexpected error, outdated information, misleading statement, insufficient context, difficulty in decision-making, lack of transparency, complex jargon, overcomplicated solution, missing critical details, irrelevant response, unconvincing argument, lack of practical application, unrealistic assumption, biased perspective, failure to address concerns, poorly structured explanation, logical fallacy, lack of citation, conflicting sources, failure to meet requirements, unanticipated consequences, incomplete analysis, ineffective troubleshooting, delayed resolution, lack of alternative solutions, misinterpretation of question, failure to adapt to context, insufficient depth, overgeneralization, misaligned priorities, oversimplified reasoning, lack of real-world examples} \\ \hline
\textbf{user\_role} & \scriptsize{user, human} \\ \hline
\textbf{assistant\_role} & \scriptsize{assistant, bot, agent} \\
\bottomrule
\end{tabular}
\caption{Placeholders and Possible Values.}
\label{tab:placeholder-values}
\end{table*}

\section{Prompts}
\label{app:prompts}

This section provides the detailed prompts used throughout our CDR benchmark development process.

\subsection{Conversation Generation Prompt}
\label{app:conv-gen-prompt}

Figure~\ref{fig:synthetic-conv-prompt} outlines the prompt for generating synthetic conversations that closely match specific queries, ensuring natural, multi-turn conversations that accurately reflect query intent while maintaining appropriate length and format. 

\subsection{Query Generation Prompt}
\label{app:query-generation-prompt}

Figure~\ref{fig:synthetic-query-prompt} shows the prompt used to generate a single synthetic search query from a set of conversations, designed to help LLMs identify key insights and patterns within conversation clusters while focusing on product management perspectives.

\subsection{Query Augmentation Prompt}
\label{app:query-aug-prompt}

Figure~\ref{fig:synthetic-query-aug-prompt} presents the prompt for augmenting the initial query by generating three hard negative examples and one alternative positive formulation, facilitating contrastive learning by creating semantically similar but functionally distinct queries. 

\subsection{Relevance Classification Prompt}
\label{app:rel-clf-prompt}

Figure~\ref{fig:relevant-classification-prompt} shows the prompt used for assessing query-conversation relevance. This prompt was used by modern LLMs during our verification process to evaluate whether conversations were relevant to specific queries.

\section{Reranker Training}
\label{app:reranker_training}

To effectively map queries to relevant conversations, we trained a specialized reranker model using approximately 300k conversations from our filtered corpus. We used LLaMa 3.3 70B \cite{llama3370b} to generate training data through a two-step process. First, we applied the synthetic query generation prompt (Figure~\ref{fig:synthetic-query-prompt}) to create one relevant query per conversation that captured the core information needs represented in the dialogue. Then, using the query augmentation prompt (Figure~\ref{fig:synthetic-query-aug-prompt}), we generated three hard negative queries (semantically similar but intentionally irrelevant) and one additional positive query (different wording but preserving intent) for each conversation.

This approach yielded approximately 1.5 million query-conversation pairs with a 2:3 positive-to-negative ratio. We fine-tuned the GTE-Multilingual-Reranker model \cite{zhang-etal-2024-mgte} using a binary cross-entropy loss function with hard negatives, a learning rate of 2e-5 with linear warmup and decay, and maximum sequence length of 8192 tokens to accommodate longer conversations. The model was trained for 3 epochs on a single NVIDIA H100 GPU.
The reranker achieved an average precision of 96.22\% on our validation set after the final epoch. We applied this model with a threshold score of 0.9 to identify candidate relevant conversations across our corpus for the final benchmark construction.

\section{Classifier Training}
\label{app:classifier_training}

We trained a specialized binary relevance classifier to validate the reliability of the mapped query-conversation relationships. This classifier was designed to distinguish relevant and irrelevant query-conversation pairs in alignment with human judgments. For training, we utilized approximately 3K relevance pairs obtained through human assessment and an additional 20K synthetic relevance pairs generated through the procedure described in \ref{app:reranker_training}. This resulted in a training dataset of approximately 23K pairs with a balanced distribution of relevant and irrelevant examples. 

Fine-tuning was performed on the Modern-BERT-base model\cite{warner2024smarterbetterfasterlonger} using a learning rate of 2e-5 with linear warm-up and decay scheduling, batch size of 128, and maximum sequence length of 8192 tokens. The model achieved an average precision of 95.2\% on the validation set. We applied this classifier, which was trained on human-verified data, to verify and filter the remaining query-conversation mappings. This ensured that only high-confidence pairs were retained in the benchmark and that relevance standards remained consistent and reliable throughout the process.

\section{Evaluation Details}
\label{app:evaluation}

\subsection{Evaluation Setup}
\label{app:evaluation_setup}

To ensure fair and consistent comparison across all evaluated models, we applied unified evaluation protocols.
Each model was tested using its original embedding dimension and maximum sequence length as specified in the official documentation. For prompt-based embedding models, we utilized the prompts without any modifications. All conversational data used in the experiments was preprocessed in a uniform manner, ensuring format consistency across all models and minimizing performance variations arising from preprocessing discrepancies.

We define three evaluation settings that differ in the granularity of the retrieval unit:

\textbf{Turn-based Evaluation:}
Each conversation turn is treated as an independent unit. For a given query, the model retrieves the most similar individual turn from the corpus. The conversation containing the retrieved turn is considered the final match.

\textbf{Sliding Chunk Evaluation (k=3):}
Conversations are segmented into overlapping chunks of three consecutive turns. Given a query, the model retrieves the most similar chunk from all chunks in the corpus. The conversation containing the retrieved chunk is selected as the final match.

\textbf{Session-based Evaluation:}
The entire conversation serves as the retrieval unit. For a given query, the model directly retrieves the most similar conversation session from the corpus.

\subsection{Efficiency Evaluation}
\label{app:evaluation_efficiency}
We also measured practical runtime metrics to evaluate real-world usability:
\begin{itemize}
\item \textbf{Ingestion Time:} The total time required to embed the entire test corpus of 9,146 conversations. This process includes tokenization, model forwarding, and storing the embeddings in memory.
\item \textbf{Inference Time:} The combined time required to: (1) embed all 1,583 queries, (2) retrieve the corresponding conversations using these embeddings, and (3) compute the final rankings. This represents the end-to-end query processing time.
\end{itemize}

All experiments were conducted with a batch size fixed at 4 for both ingestion and inference measurements. The reported times represent the total elapsed time for processing the entire dataset.

\subsection{Hardware Specifications}
\label{app:evaluation_spec}
All evaluations were conducted under the same setup, and the hardware specifications are summarized in Table~\ref{tab:hardware-specs}.

\begin{table}[htbp]
\centering
\small
\begin{tabular}{l|l}
\toprule
\textbf{Component} & \textbf{Specification} \\
\midrule
CPU & Intel(R) Xeon(R) Platinum 8468 \\
GPU & NVIDIA H100 80GB HBM3 \\
Memory & 206GB RAM \\
\bottomrule
\end{tabular}
\caption{Hardware specifications used for all experimental evaluations.}
\label{tab:hardware-specs}
\end{table}

\section{Additional Experimental Results}

\subsection{Performance by Additional Metrics}
\label{apdx:additional_metric}

To provide a comprehensive evaluation beyond the primary results reported in Table~\ref{table:embedding-model-result}, we present an extensive analysis of model performance across multiple evaluation metrics. We evaluate all models under three distinct retrieval configurations: Turn-based, Sliding Chunk (k=3), and Session-based approaches. For each configuration, we report performance across five key metrics: Accuracy (ACC), Precision (P), Recall (R), Normalized Discounted Cumulative Gain (NDCG), and Mean Reciprocal Rank (MRR), evaluated at cutoff thresholds of 1, 5, 10, and 20. The detailed results are systematically presented in Tables~\ref{table:turn-based-model-additional-result}, \ref{table:sliding-chunk-model-additional-result}, and \ref{table:session-based-model-additional-result}, respectively. This multi-faceted evaluation framework enables a thorough assessment of model efficacy across varying retrieval granularities and ranking depths, providing deeper insights into the comparative strengths and limitations of each approach.

\begin{table*}[thbp]
    \centering
    \begin{adjustbox}{max width=\textwidth}
    \renewcommand{\arraystretch}{1.1}
    \begin{tabular}{l c c c c c c c c c c c c c c c c c c c c}
        \toprule[1.2pt]
        \multirow{2}{*}{\textbf{Model}} & \multicolumn{20}{c}{\textbf{Turn}} \\
        \cmidrule(lr){2-21}
        \multicolumn{1}{c}{} & \textbf{ACC@1} & \textbf{ACC@5} & \textbf{ACC@10} & \textbf{ACC@20} & \textbf{P@1} & \textbf{P@5} & \textbf{P@10} & \textbf{P@20} & \textbf{R@1} & \textbf{R@5} & \textbf{R@10} & \textbf{R@20} & \textbf{NDCG@1} & \textbf{NDCG@5} & \textbf{NDCG@10} & \textbf{NDCG@20} & \textbf{MRR@1} & \textbf{MRR@5} & \textbf{MRR@10} & \textbf{MRR@20} \\
        \midrule[1pt]
        \rowcolor{gray!15} \multicolumn{21}{l}{\textbf{Commercial API Models}} \\
        \cmidrule[0.8pt]{1-21}
        Voyage-3-large & 0.6816 & 0.8440 & 0.8876 & 0.9210 & 0.6816 & 0.5495 & 0.4359 & 0.3139 & 0.0433 & 0.1727 & 0.2609 & 0.3549 & 0.6816 & 0.5828 & 0.5079 & 0.4530 & 0.6816 & 0.7470 & 0.7530 & 0.7553 \\
        \midrule[0.3pt]
        Text-embedding-3-large & 0.6684 & 0.8446 & 0.9033 & 0.9394 & 0.6684 & 0.5453 & 0.4389 & 0.3137 & 0.0444 & 0.1758 & 0.2698 & 0.3638 & 0.6684 & 0.5763 & 0.5078 & 0.4552 & 0.6684 & 0.7384 & 0.7463 & 0.7488 \\
        Text-embedding-3-small & 0.6601 & 0.8547 & 0.9059 & 0.9394 & 0.6601 & 0.5325 & 0.4183 & 0.3031 & 0.0433 & 0.1708 & 0.2558 & 0.3508 & 0.6601 & 0.5652 & 0.4897 & 0.4415 & 0.6601 & 0.7383 & 0.7451 & 0.7475 \\
        \midrule[0.3pt]
        Embed-english-v3.0 & 0.5799 & 0.7991 & 0.8642 & 0.9135 & 0.5799 & 0.4550 & 0.3547 & 0.2506 & 0.0382 & 0.1503 & 0.2237 & 0.3015 & 0.5799 & 0.4847 & 0.4189 & 0.3760 & 0.5799 & 0.6662 & 0.6750 & 0.6783 \\
        \midrule[1pt]
        \rowcolor{gray!15} \multicolumn{21}{l}{\textbf{Open Source Models}} \\
        \cmidrule[0.8pt]{1-21}
        Stella\_en\_1.5B\_v5 & 0.6759 & 0.8421 & 0.8926 & 0.9324 & 0.6759 & 0.5320 & 0.4141 & 0.2900 & 0.0453 & 0.1776 & 0.2592 & 0.3443 & 0.6759 & 0.5688 & 0.4907 & 0.4365 & 0.6759 & 0.7424 & 0.7495 & 0.7523 \\
        Stella\_en\_400M\_v5 & 0.6450 & 0.8313 & 0.8863 & 0.9191 & 0.6450 & 0.5051 & 0.3963 & 0.2794 & 0.0424 & 0.1675 & 0.2490 & 0.3338 & 0.6450 & 0.5402 & 0.4682 & 0.4187 & 0.6450 & 0.7170 & 0.7245 & 0.7269 \\
        \midrule[0.3pt]
        Jasper\_en\_vision\_language\_v1 & 0.6045 & 0.7922 & 0.8515 & 0.8989 & 0.6045 & 0.4710 & 0.3712 & 0.2643 & 0.0406 & 0.1557 & 0.2317 & 0.3117 & 0.6045 & 0.5042 & 0.4379 & 0.3927 & 0.6045 & 0.6749 & 0.6826 & 0.6860 \\
        \midrule[0.3pt]
        NV-Embed-v2 & 0.0013 & 0.7410 & 0.8105 & 0.8617 & 0.0013 & 0.3805 & 0.3251 & 0.2310 & 0.0000 & 0.1230 & 0.2008 & 0.2707 & 0.0013 & 0.3258 & 0.3170 & 0.2929 & 0.0013 & 0.3398 & 0.3492 & 0.3528 \\
        NV-Embed-v1 & 0.4062 & 0.6008 & 0.6721 & 0.7473 & 0.4062 & 0.2678 & 0.1956 & 0.1338 & 0.0279 & 0.0871 & 0.1226 & 0.1618 & 0.4062 & 0.2999 & 0.2467 & 0.2165 & 0.4062 & 0.4791 & 0.4888 & 0.4941 \\
        \midrule[0.3pt]
        SFR-Embedding-2\_R & 0.4864 & 0.7012 & 0.7732 & 0.8326 & 0.4864 & 0.3588 & 0.2805 & 0.1980 & 0.0326 & 0.1189 & 0.1775 & 0.2393 & 0.4864 & 0.3881 & 0.3344 & 0.3003 & 0.4864 & 0.5684 & 0.5778 & 0.5821 \\
        \midrule[0.3pt]
        Jina-embeddings-v3 & 0.5502 & 0.7656 & 0.8200 & 0.8585 & 0.5502 & 0.4129 & 0.3160 & 0.2271 & 0.0381 & 0.1408 & 0.2053 & 0.2792 & 0.5502 & 0.4449 & 0.3803 & 0.3457 & 0.5502 & 0.6316 & 0.6389 & 0.6416 \\
        \midrule[0.3pt]
        Modernbert-embed-base & 0.5066 & 0.7505 & 0.8313 & 0.8844 & 0.5066 & 0.3821 & 0.3026 & 0.2185 & 0.0353 & 0.1269 & 0.1923 & 0.2650 & 0.5066 & 0.4125 & 0.3594 & 0.3269 & 0.5066 & 0.5988 & 0.6098 & 0.6136 \\
        \midrule[0.3pt]
        Gte-Qwen2-1.5B-instruct & 0.6387 & 0.8225 & 0.8749 & 0.9128 & 0.6387 & 0.5054 & 0.3952 & 0.2798 & 0.0419 & 0.1607 & 0.2412 & 0.3233 & 0.6387 & 0.5380 & 0.4646 & 0.4127 & 0.6387 & 0.7102 & 0.7171 & 0.7199 \\
        Gte-large-en-v1.5 & 0.4668 & 0.6961 & 0.7732 & 0.8244 & 0.4668 & 0.3522 & 0.2792 & 0.2007 & 0.0319 & 0.1186 & 0.1821 & 0.2506 & 0.4668 & 0.3788 & 0.3310 & 0.3025 & 0.4668 & 0.5517 & 0.5620 & 0.5655 \\
        \midrule[0.3pt]
        Bge-large-en-v1.5 & 0.4738 & 0.7202 & 0.7833 & 0.8345 & 0.4738 & 0.3554 & 0.2719 & 0.1932 & 0.0328 & 0.1208 & 0.1757 & 0.2383 & 0.4738 & 0.3836 & 0.3276 & 0.2963 & 0.4738 & 0.5663 & 0.5749 & 0.5786 \\
        \midrule[0.3pt]
        Cde-small-v2 & 0.1718 & 0.3424 & 0.4195 & 0.5205 & 0.1718 & 0.1265 & 0.0975 & 0.0758 & 0.0114 & 0.0403 & 0.0606 & 0.0911 & 0.1718 & 0.1370 & 0.1163 & 0.1106 & 0.1718 & 0.2328 & 0.2430 & 0.2501 \\
        \bottomrule[1.2pt]
    \end{tabular}
    \end{adjustbox}
    \caption{Additional performance metrics for turn-based evaluation setting.}
    \label{table:turn-based-model-additional-result}
\end{table*} 

\begin{table*}[thbp]
    \centering
    \begin{adjustbox}{max width=\textwidth}
    \renewcommand{\arraystretch}{1.1}
    \begin{tabular}{l c c c c c c c c c c c c c c c c c c c c}
        \toprule[1.2pt]
        \multirow{2}{*}{\textbf{Model}} & \multicolumn{20}{c}{\textbf{Sliding chunk (k=3)}} \\
        \cmidrule(lr){2-21}
        \multicolumn{1}{c}{} & \textbf{ACC@1} & \textbf{ACC@5} & \textbf{ACC@10} & \textbf{ACC@20} & \textbf{P@1} & \textbf{P@5} & \textbf{P@10} & \textbf{P@20} & \textbf{R@1} & \textbf{R@5} & \textbf{R@10} & \textbf{R@20} & \textbf{NDCG@1} & \textbf{NDCG@5} & \textbf{NDCG@10} & \textbf{NDCG@20} & \textbf{MRR@1} & \textbf{MRR@5} & \textbf{MRR@10} & \textbf{MRR@20} \\
        \midrule[1pt]
        \rowcolor{gray!15} \multicolumn{21}{l}{\textbf{Commercial API Models}} \\
        \cmidrule[0.8pt]{1-21}
        Voyage-3-large & 0.6860 & 0.8440 & 0.8857 & 0.9286 & 0.6860 & 0.5497 & 0.4327 & 0.3167 & 0.0445 & 0.1739 & 0.2582 & 0.3579 & 0.6860 & 0.5844 & 0.5063 & 0.4557 & 0.6860 & 0.7478 & 0.7535 & 0.7565 \\
        \midrule[0.3pt]
        Text-embedding-3-large & 0.6810 & 0.8440 & 0.8977 & 0.9400 & 0.6810 & 0.5568 & 0.4416 & 0.3170 & 0.0442 & 0.1785 & 0.2696 & 0.3666 & 0.6810 & 0.5881 & 0.5130 & 0.4599 & 0.6810 & 0.7439 & 0.7514 & 0.7544 \\
        Text-embedding-3-small & 0.6747 & 0.8408 & 0.8920 & 0.9343 & 0.6747 & 0.5289 & 0.4171 & 0.2990 & 0.0440 & 0.1669 & 0.2528 & 0.3462 & 0.6747 & 0.5644 & 0.4894 & 0.4381 & 0.6747 & 0.7401 & 0.7471 & 0.7501 \\ 
        \midrule[0.3pt]
        Embed-english-v3.0 & 0.3721 & 0.6090 & 0.6772 & 0.7612 & 0.3721 & 0.2786 & 0.2116 & 0.1492 & 0.0255 & 0.0909 & 0.1351 & 0.1824 & 0.3721 & 0.2998 & 0.2547 & 0.2290 & 0.3721 & 0.4603 & 0.4696 & 0.4755 \\
        \midrule[1pt]
        \rowcolor{gray!15} \multicolumn{21}{l}{\textbf{Open Source Models}} \\
        \cmidrule[0.8pt]{1-21}
        Stella\_en\_1.5B\_v5 & 0.6740 & 0.8446 & 0.8920 & 0.9324 & 0.6740 & 0.5301 & 0.4078 & 0.2872 & 0.0442 & 0.1739 & 0.2558 & 0.3396 & 0.6740 & 0.5663 & 0.4855 & 0.4320 & 0.6740 & 0.7405 & 0.7469 & 0.7497 \\
        Stella\_en\_400M\_v5 & 0.6450 & 0.8219 & 0.8730 & 0.9135 & 0.6450 & 0.5039 & 0.3919 & 0.2732 & 0.0425 & 0.1671 & 0.2462 & 0.3268 & 0.6450 & 0.5393 & 0.4651 & 0.4127 & 0.6450 & 0.7143 & 0.7214 & 0.7242 \\
        \midrule[0.3pt]
        Jasper\_en\_vision\_language\_v1 & 0.6172 & 0.7802 & 0.8326 & 0.8768 & 0.6172 & 0.4667 & 0.3615 & 0.2498 & 0.0406 & 0.1513 & 0.2245 & 0.2951 & 0.6172 & 0.5024 & 0.4309 & 0.3793 & 0.6172 & 0.6779 & 0.6851 & 0.6883 \\
        \midrule[0.3pt]
        NV-Embed-v2 & 0.5970 & 0.7410 & 0.7890 & 0.8345 & 0.5970 & 0.4335 & 0.3262 & 0.2237 & 0.0391 & 0.1378 & 0.1988 & 0.2610 & 0.5970 & 0.4709 & 0.3956 & 0.3452 & 0.5970 & 0.6530 & 0.6596 & 0.6629 \\
        NV-Embed-v1 & 0.4195 & 0.5989 & 0.6557 & 0.7233 & 0.4195 & 0.2870 & 0.2080 & 0.1392 & 0.0282 & 0.0926 & 0.1302 & 0.1702 & 0.4195 & 0.3178 & 0.2603 & 0.2262 & 0.4195 & 0.4903 & 0.4980 & 0.5027 \\
        \midrule[0.3pt]
        SFR-Embedding-2\_R & 0.4630 & 0.6526 & 0.7075 & 0.7625 & 0.4630 & 0.3395 & 0.2589 & 0.1790 & 0.0315 & 0.1121 & 0.1639 & 0.2186 & 0.4630 & 0.3680 & 0.3127 & 0.2778 & 0.4630 & 0.5370 & 0.5445 & 0.5480 \\
        \midrule[0.3pt]
        Jina-embeddings-v3 & 0.5483 & 0.7612 & 0.8212 & 0.8737 & 0.5483 & 0.4312 & 0.3363 & 0.2423 & 0.0382 & 0.1443 & 0.2142 & 0.2919 & 0.5483 & 0.4607 & 0.3983 & 0.3618 & 0.5483 & 0.6322 & 0.6403 & 0.6439 \\
        \midrule[0.3pt]
        Modernbert-embed-base & 0.4839 & 0.7176 & 0.7928 & 0.8471 & 0.4839 & 0.3645 & 0.2857 & 0.2062 & 0.0330 & 0.1206 & 0.1795 & 0.2501 & 0.4839 & 0.3927 & 0.3398 & 0.3090 & 0.4839 & 0.5706 & 0.5806 & 0.5845 \\
        \midrule[0.3pt]
        Gte-Qwen2-1.5B-instruct & 0.6159 & 0.7751 & 0.8395 & 0.8863 & 0.6159 & 0.4738 & 0.3708 & 0.2602 & 0.0397 & 0.1513 & 0.2261 & 0.3042 & 0.6159 & 0.5087 & 0.4386 & 0.3882 & 0.6159 & 0.6792 & 0.6876 & 0.6908 \\
        Gte-large-en-v1.5 & 0.4548 & 0.6721 & 0.7505 & 0.8193 & 0.4548 & 0.3479 & 0.2726 & 0.1923 & 0.0313 & 0.1186 & 0.1778 & 0.2408 & 0.4548 & 0.3741 & 0.3246 & 0.2936 & 0.4548 & 0.5379 & 0.5485 & 0.5534 \\
        \midrule[0.3pt]
        Bge-large-en-v1.5 & 0.4618 & 0.6810 & 0.7612 & 0.8143 & 0.4618 & 0.3332 & 0.2539 & 0.1772 & 0.0330 & 0.1138 & 0.1659 & 0.2204 & 0.4618 & 0.3643 & 0.3105 & 0.2780 & 0.4618 & 0.5419 & 0.5528 & 0.5564 \\
        \midrule[0.3pt]
        Cde-small-v2 & 0.1876 & 0.3392 & 0.4188 & 0.5155 & 0.1876 & 0.1318 & 0.1007 & 0.0764 & 0.0129 & 0.0442 & 0.0640 & 0.0923 & 0.1876 & 0.1450 & 0.1226 & 0.1148 & 0.1876 & 0.2425 & 0.2531 & 0.2600 \\
        \bottomrule[1.2pt]
    \end{tabular}
    \end{adjustbox}
    \caption{Additional performance metrics for sliding chunk (k=3) evaluation setting.}
    \label{table:sliding-chunk-model-additional-result}
\end{table*} 

\begin{table*}[thbp]
    \centering
    \begin{adjustbox}{max width=\textwidth}
    \renewcommand{\arraystretch}{1.1}
    \begin{tabular}{l c c c c c c c c c c c c c c c c c c c c}
        \toprule[1.2pt]
        \multirow{2}{*}{\textbf{Model}} & \multicolumn{20}{c}{\textbf{Session}} \\
        \cmidrule(lr){2-21}
        \multicolumn{1}{c}{} & \textbf{ACC@1} & \textbf{ACC@5} & \textbf{ACC@10} & \textbf{ACC@20} & \textbf{P@1} & \textbf{P@5} & \textbf{P@10} & \textbf{P@20} & \textbf{R@1} & \textbf{R@5} & \textbf{R@10} & \textbf{R@20} & \textbf{NDCG@1} & \textbf{NDCG@5} & \textbf{NDCG@10} & \textbf{NDCG@20} & \textbf{MRR@1} & \textbf{MRR@5} & \textbf{MRR@10} & \textbf{MRR@20} \\
        \midrule[1pt]
        \rowcolor{gray!15} \multicolumn{21}{l}{\textbf{Commercial API Models}} \\
        \cmidrule[0.8pt]{1-21}
        Voyage-3-large & 0.6639 & 0.8294 & 0.8844 & 0.9198 & 0.6639 & 0.5381 & 0.4358 & 0.3158 & 0.0436 & 0.1686 & 0.2615 & 0.3564 & 0.6639 & 0.5700 & 0.5036 & 0.4521 & 0.6639 & 0.7317 & 0.7392 & 0.7417 \\
        \midrule[0.3pt]
        Text-embedding-3-large & 0.6513 & 0.8174 & 0.8838 & 0.9280 & 0.6513 & 0.5291 & 0.4190 & 0.2977 & 0.0410 & 0.1661 & 0.2529 & 0.3435 & 0.6513 & 0.5604 & 0.4876 & 0.4340 & 0.6513 & 0.7160 & 0.7249 & 0.7281 \\
        Text-embedding-3-small &  0.6431 & 0.8174 & 0.8787 & 0.9179 & 0.6431 & 0.5041 & 0.3972 & 0.2829 & 0.0414 & 0.1608 & 0.2412 & 0.3266 & 0.6431 & 0.5385 & 0.4664 & 0.4161 & 0.6431 & 0.7118 & 0.7199 & 0.7227 \\
        \midrule[0.3pt]
        Embed-english-v3.0 & 0.5224 & 0.7309 & 0.7953 & 0.8528 & 0.5224 & 0.3960 & 0.2987 & 0.2070 & 0.0356 & 0.1352 & 0.1923 & 0.2530 & 0.5224 & 0.4271 & 0.3620 & 0.3223 & 0.5224 & 0.6027 & 0.6115 & 0.6156 \\
        \midrule[1pt]
        \rowcolor{gray!15} \multicolumn{21}{l}{\textbf{Open Source Models}} \\
        \cmidrule[0.8pt]{1-21}
        Stella\_en\_1.5B\_v5 & 0.6627 & 0.8269 & 0.8762 & 0.9154 & 0.6627 & 0.5119 & 0.3961 & 0.2740 & 0.0440 & 0.1682 & 0.2481 & 0.3259 & 0.6627 & 0.5493 & 0.4722 & 0.4162 & 0.6627 & 0.7258 & 0.7322 & 0.7349 \\
        Stella\_en\_400M\_v5 & 0.6374 & 0.8130 & 0.8636 & 0.9046 & 0.6374 & 0.5003 & 0.3846 & 0.2684 & 0.0426 & 0.1650 & 0.2400 & 0.3188 & 0.6374 & 0.5352 & 0.4583 & 0.4062 & 0.6374 & 0.7054 & 0.7121 & 0.7150 \\
        \midrule[0.3pt]
        Jasper\_en\_vision\_language\_v1 & 0.6406 & 0.7997 & 0.8547 & 0.9015 & 0.6406 & 0.4968 & 0.3814 & 0.2635 & 0.0430 & 0.1626 & 0.2382 & 0.3131 & 0.6406 & 0.5334 & 0.4561 & 0.4017 & 0.6406 & 0.7039 & 0.7116 & 0.7147 \\
        \midrule[0.3pt]
        NV-Embed-v2 & 0.6500 & 0.8042 & 0.8490 & 0.8913 & 0.6500 & 0.5040 & 0.3855 & 0.2638 & 0.0420 & 0.1606 & 0.2344 & 0.3071 & 0.6500 & 0.5395 & 0.4592 & 0.4002 & 0.6500 & 0.7098 & 0.7159 & 0.7187 \\
        NV-Embed-v1 & 0.6380 & 0.8105 & 0.8591 & 0.9015 & 0.6380 & 0.4787 & 0.3634 & 0.2464 & 0.0418 & 0.1547 & 0.2242 & 0.2894 & 0.6380 & 0.5184 & 0.4389 & 0.3804 & 0.6380 & 0.7034 & 0.7101 & 0.7131 \\
        \midrule[0.3pt]
        SFR-Embedding-2\_R & 0.6393 & 0.7890 & 0.8250 & 0.8642 & 0.6393 & 0.4931 & 0.3722 & 0.2534 & 0.0416 & 0.1584 & 0.2280 & 0.2970 & 0.6393 & 0.5293 & 0.4474 & 0.3889 & 0.6393 & 0.6973 & 0.7022 & 0.7049 \\
        \midrule[0.3pt]
        Jina-embeddings-v3 & 0.5287 & 0.7366 & 0.7997 & 0.8503 & 0.5287 & 0.4030 & 0.3088 & 0.2182 & 0.0361 & 0.1369 & 0.1995 & 0.2665 & 0.5287 & 0.4345 & 0.3718 & 0.3347 & 0.5287 & 0.6079 & 0.6165 & 0.6200 \\
        \midrule[0.3pt]
        Modernbert-embed-base & 0.5060 & 0.7239 & 0.7922 & 0.8484 & 0.5060 & 0.3827 & 0.3016 & 0.2129 & 0.0353 & 0.1258 & 0.1906 & 0.2578 & 0.5060 & 0.4121 & 0.3579 & 0.3217 & 0.5060 & 0.5900 & 0.5993 & 0.6033 \\
        \midrule[0.3pt]
        Gte-Qwen2-1.5B-instruct & 0.5268 & 0.7404 & 0.8099 & 0.8598 & 0.5268 & 0.3895 & 0.2987 & 0.2113 & 0.0379 & 0.1295 & 0.1919 & 0.2585 & 0.5268 & 0.4226 & 0.3615 & 0.3250 & 0.5268 & 0.6099 & 0.6192 & 0.6227 \\
        Gte-large-en-v1.5 & 0.4877 & 0.6949 & 0.7694 & 0.8244 & 0.4877 & 0.3665 & 0.2860 & 0.1997 & 0.0334 & 0.1232 & 0.1840 & 0.2435 & 0.4877 & 0.3972 & 0.3429 & 0.3064 & 0.4877 & 0.5688 & 0.5789 & 0.5828 \\
        \midrule[0.3pt]
        Bge-large-en-v1.5 & 0.4769 & 0.6961 & 0.7618 & 0.8256 & 0.4769 & 0.3349 & 0.2476 & 0.1732 & 0.0333 & 0.1148 & 0.1617 & 0.2164 & 0.4769 & 0.3674 & 0.3071 & 0.2755 & 0.4769 & 0.5590 & 0.5678 & 0.5722 \\
        \midrule[0.3pt]
        Cde-small-v2 & 0.1175 & 0.2527 & 0.3392 & 0.4251 & 0.1175 & 0.0853 & 0.0701 & 0.0521 & 0.0091 & 0.0292 & 0.0463 & 0.0683 & 0.1175 & 0.0932 & 0.0830 & 0.0792 & 0.1175 & 0.1660 & 0.1776 & 0.1836 \\
        \bottomrule[1.2pt]
    \end{tabular}
    \end{adjustbox}
    \caption{Additional performance metrics for session-based evaluation setting.}
    \label{table:session-based-model-additional-result}
\end{table*}

\subsection{Performance per Tasks}
\label{apdx:all_result}
Table~\ref{table:embedding-model-result-per-task} presents the detailed NDCG@10, Recall@10, and Precision@10 performance of all evaluated embedding models across the five task categories in our benchmark. As illustrated in Figure~\ref{fig:result-per-task}, performance varies significantly between task types, with most models showing strengths in content-oriented categories like `Emotion \& Feedback' and `Intent \& Purpose' while struggling with interaction-focused categories, particularly `Conversation Dynamics'.

The table highlights the lack of a universally dominant approach for conversational data retrieval tasks. Even top-performing models like Voyage-3-large demonstrate inconsistent performance across different categories. Notably, `Conversation Dynamics' remains challenging for all models, with the highest scores barely reaching 0.17, indicating a substantial opportunity for architectural improvements specifically designed to capture conversation flow and structure.

\begin{table*}[htbp]
    \centering
    \begin{adjustbox}{max width=\textwidth}
    \renewcommand{\arraystretch}{1.2}
    \begin{tabular}{l c c c c c c c c c c c c c c c}
        \toprule[1.2pt]
        \multirow{3}{*}{\textbf{Model}} & \multicolumn{3}{c}{\textbf{\makecell{Conversation\\Dynamics}}} & \multicolumn{3}{c}{\textbf{\makecell{Emotion\\\&\\Feedback}}} & \multicolumn{3}{c}{\textbf{\makecell{Intent\\\&\\Purpose}}} & \multicolumn{3}{c}{\textbf{\makecell{Linguistic Style\\\&\\Expression}}} & \multicolumn{3}{c}{\textbf{\makecell{Trust, Safety\\\&\\Ethics}}} \\
        \cmidrule(lr){2-4} \cmidrule(lr){5-7} \cmidrule(lr){8-10} \cmidrule(lr){11-13} \cmidrule(lr){14-16}
        & \textbf{NDCG@10} & \textbf{R@10} & \textbf{P@10} & \textbf{NDCG@10} & \textbf{R@10} & \textbf{P@10} & \textbf{NDCG@10} & \textbf{R@10} & \textbf{P@10} & \textbf{NDCG@10} & \textbf{R@10} & \textbf{P@10} & \textbf{NDCG@10} & \textbf{R@10} & \textbf{P@10} \\
        \midrule[1pt]
        \rowcolor{gray!15} \multicolumn{16}{l}{\textbf{Commercial API Models}} \\
        \cmidrule[0.8pt]{1-16}
        Voyage-3-large \cite{voyage3large} & 0.0930 & 0.0466 & 0.0825 & \textbf{0.6394} & \textbf{0.3972} & \textbf{0.5289} & \textbf{0.6211} & \textbf{0.2956} & \textbf{0.5604} & \textbf{0.3373} & \textbf{0.1109} & \textbf{0.2806} & \underline{0.5628} & \textbf{0.2689} & \underline{0.4974} \\
        \cmidrule[0.5pt]{1-16}
        Text-embedding-3-large \cite{openaiembedding} & 0.1124 & 0.0576 & 0.1014 & \underline{0.6368} & \underline{0.3849} & \underline{0.5214} & \underline{0.5919} & \underline{0.2829} & \underline{0.5237} & \underline{0.2896} & \underline{0.0934} & \underline{0.2456} & \textbf{0.5631} & \underline{0.2617} & \textbf{0.4991} \\
        Text-embedding-3-small \cite{openaiembedding} & \textbf{0.1386} & \textbf{0.0663} & \textbf{0.1199} & 0.5977 & 0.3637 & 0.4865 & 0.5644 & 0.2672 & 0.5010 & 0.2655 & 0.0836 & 0.2151 & 0.5461 & 0.2580 & 0.4788 \\
        \cmidrule[0.5pt]{1-16}
        Embed-english-v3.0 \cite{embedv3} & \underline{0.1339} & \underline{0.0651} & \underline{0.1133} & 0.5649 & 0.3434 & 0.4535 & 0.3883 & 0.1851 & 0.3349 & 0.1966 & 0.0644 & 0.1504 & 0.3926 & 0.1837 & 0.3355 \\
        \midrule[1pt]
        \rowcolor{gray!15} \multicolumn{16}{l}{\textbf{Open Source Models}} \\
        \cmidrule[0.8pt]{1-16}
        Stella\_en\_1.5B\_v5 \cite{zhang2025jasperstelladistillationsota} & 0.1409 & 0.0682 & 0.1161 & 0.6586 & 0.4019 & 0.5390 & \textbf{0.5341} & \textbf{0.2568} & \textbf{0.4679} & 0.2972 & 0.0979 & 0.2369 & \underline{0.5366} & \underline{0.2516} & \underline{0.4597} \\
        Stella\_en\_400M\_v5 \cite{zhang2025jasperstelladistillationsota} & 0.1390 & 0.0682 & 0.1171 & 0.6519 & 0.3943 & 0.5314 & \underline{0.5162} & \underline{0.2463} & \underline{0.4527} & 0.2599 & 0.0848 & 0.2032 & \textbf{0.5406} & \textbf{0.2552} & \textbf{0.4662} \\
        \cmidrule[0.5pt]{1-16}
        Jasper\_en\_vision\_language\_v1 \cite{zhang2025jasperstelladistillationsota} & 0.1134 & 0.0550 & 0.0991 & \underline{0.6645} & \underline{0.4061} & \underline{0.5462} & 0.5099 & 0.2428 & 0.4443 & 0.2841 & 0.0931 & 0.2226 & 0.5174 & 0.2354 & 0.4355 \\
        \cmidrule[0.5pt]{1-16}
        NV-Embed-v2 \cite{lee2024nvembed} & 0.0846 & 0.0374 & 0.0659 & \textbf{0.6797} & \textbf{0.4185} & \textbf{0.5660} & 0.4955 & 0.2239 & 0.4293 & \underline{0.3242} & \underline{0.1066} & \underline{0.2655} & 0.5341 & 0.2439 & 0.4571 \\
        NV-Embed-v1 \cite{lee2024nvembed} & 0.1280 & 0.0533 & 0.1047 & 0.6355 & 0.3830 & 0.5198 & 0.4966 & 0.2314 & 0.4297 & 0.3001 & 0.1004 & 0.2448 & 0.4446 & 0.1988 & 0.3619 \\
        \cmidrule[0.5pt]{1-16}
        SFR-Embedding-2\_R \cite{SFRAIResearch2024} & 0.0752 & 0.0318 & 0.0597 & 0.6383 & 0.3881 & 0.5151 & 0.5100 & 0.2391 & 0.4450 & \textbf{0.3521} & \textbf{0.1097} & \textbf{0.2849} & 0.4686 & 0.2226 & 0.3918 \\
        \cmidrule[0.5pt]{1-16}
        Jina-embeddings-v3 \cite{sturua2024jinaembeddingsv3multilingualembeddingstask} & 0.1416 & \underline{0.0712} & 0.1209 & 0.5942 & 0.3641 & 0.4843 & 0.3852 & 0.1866 & 0.3349 & 0.2062 & 0.0701 & 0.1615 & 0.4166 & 0.1932 & 0.3524 \\
        \cmidrule[0.5pt]{1-16}
        Modernbert-embed-base \cite{warner2024smarterbetterfasterlonger} & \textbf{0.1848} & \textbf{0.0854} & \textbf{0.1512} & 0.5479 & 0.3339 & 0.4403 & 0.3950 & 0.1907 & 0.3571 & 0.1702 & 0.0576 & 0.1341 & 0.3700 & 0.1736 & 0.3182 \\
        \cmidrule[0.5pt]{1-16}
        Gte-Qwen2-1.5B-instruct \cite{li2023generaltextembeddingsmultistage} & 0.1270 & 0.0550 & 0.1081 & 0.5190 & 0.3206 & 0.4208 & 0.3774 & 0.1799 & 0.3251 & 0.2451 & 0.0837 & 0.1984 & 0.3248 & 0.1576 & 0.2740 \\
        Gte-large-en-v1.5 \cite{li2023generaltextembeddingsmultistage} & 0.0715 & 0.0329 & 0.0602 & 0.5806 & 0.3606 & 0.4711 & 0.3407 & 0.1619 & 0.3016 & 0.2183 & 0.0748 & 0.1770 & 0.3861 & 0.1813 & 0.3208 \\
        \cmidrule[0.5pt]{1-16}
        Bge-large-en-v1.5 \cite{bge_embedding} & \underline{0.1486} & 0.0693 & \underline{0.1242} & 0.5033 & 0.3058 & 0.4006 & 0.3044 & 0.1441 & 0.2576 & 0.1959 & 0.0608 & 0.1488 & 0.2996 & 0.1363 & 0.2433 \\
        \cmidrule[0.5pt]{1-16}
        Cde-small-v2 \cite{morris2025contextual} & 0.0703 & 0.0347 & 0.0626 & 0.1333 & 0.0847 & 0.1066 & 0.086 & 0.0422 & 0.0778 & 0.048 & 0.0148 & 0.0389 & 0.0576 & 0.0267 & 0.0485 \\
        \bottomrule[1.2pt]
    \end{tabular}
    \end{adjustbox}
    \caption{This table presents task-specific NDCG@10, Recall@10 (R@10), and Precision@10 (P@10) scores for all evaluated embedding models across the five task categories of the CDR benchmark. The results highlight consistent performance patterns across categories, with most models showing strongest performance on Emotion \& Feedback and weakest performance on Conversation Dynamics.}
    \label{table:embedding-model-result-per-task}
\end{table*}

\section{Dataset License and Disclaimer}
\label{app:license}

In this work, we utilize multiple publicly available open-source dialogue datasets to construct our initial data pool. The LMSYS-Chat-1M dataset is distributed under a custom LMSYS-Chat-1M License Agreement and is non-redistributable. The WildChat-1M-Full dataset is licensed under ODC-BY 1.0 (Open Data Commons Attribution). The Bitext Customer Support dataset is released under the CDLA-Sharing 1.0 license. The Schema-Guided Dialogue (SGD) dataset is provided under a CC BY-SA 4.0 license. The DialogSum and DailyDialog datasets are released under CC BY-NC-SA 4.0, while OpenDialKG and Bot-Adversarial Dialogue (BAD) are licensed under CC BY-NC 4.0, with the latter restricted to research use only. The MultiWOZ 2.2 and SmolTalk datasets are licensed under Apache 2.0. The MobileConvRec dataset is released under a CC BY 4.0 license. All datasets were used strictly for research purposes in compliance with their respective licenses. No personally identifiable information (PII) was included, and all data was either anonymized by the original providers or manually verified during the curation process. We do not claim ownership of these datasets and respectfully acknowledge the contributions of their original authors and publishers.

\begin{figure*}[!ht]
\centering
\begin{tcolorbox}
\begin{spacing}{1.0}
\scriptsize{
You are an AI assistant specializing in adapting conversations to match specific search queries. Your task is to modify a given reference conversation so that it appears highly relevant to a provided search query while maintaining a natural tone and semantic similarity.\\
\\
Here are the inputs for this task:\\
\\
Reference Conversation:\\
<reference\_conversation>\\
\{conversation\}\\
</reference\_conversation>\\
\\
Search Query:\\
<search\_query>\\
\{query\}\\
</search\_query>\\
\\
Instructions:\\
\\
1. Analyze the search query thoroughly to understand its content, sentiment, and any temporal aspects (e.g., changes in emotion or situation over time).\\
\\
2. Review the reference conversation and identify key elements that can be modified to align with the search query.\\
\\
3. Develop an adaptation strategy by wrapping your analysis in <adaptation\_strategy> tags. Follow these steps:\\
\hspace*{2em}   a. Query analysis: List and number key concepts, sentiment, and temporal aspects\\
\hspace*{2em}   b. Reference conversation analysis: List and number key elements and themes\\
\hspace*{2em}   c. Similarity mapping: Match query elements to conversation parts\\
\hspace*{2em}   d. Adaptation planning: Outline specific changes to be made\\
\hspace*{2em}   e. Emotion and tone alignment: Ensure the adapted conversation matches the query's sentiment\\
\hspace*{2em}   f. Natural language preservation: Plan how to maintain a natural flow\\
\hspace*{2em}   g. Adaptation feasibility: Explicitly state whether adaptation is possible or if a ``REJECT'' output is necessary\\
\hspace*{2em}   h. Turn count analysis: Ensure the adapted conversation has a similar number of turns as the original\\
\hspace*{2em}   It's okay for this section to be quite long, as it involves detailed analysis and planning.\\
\\
4. Based on your adaptation strategy, modify the conversation to make it semantically relevant to the search query. Ensure that:\\
\hspace*{2em}   - The conversation expresses the situation or sentiment described in the query clearly and unmistakably\\
\hspace*{2em}   - Avoid overly literal translations or excessive keyword matching\\
\hspace*{2em}   - The adapted conversation has a similar number of turns as the original\\
\hspace*{2em}   - The conversation is always multi-turn (at least two exchanges between user and assistant)\\
\\
5. If the query indicates a change over time (e.g., ``initially liked something but became angry later''), reflect this progression in the adapted conversation.\\
\\
6. Maintain a natural tone similar to the reference conversation while prioritizing the accurate representation of the query's sentiment.\\
\\
7. Use only ``user'' and ``assistant'' roles in the adapted conversation.\\
\\
8. If the adaptation is not possible or would result in an unnatural conversation, output only the string ``REJECT'' and end the process.\\
\\
9. If adaptation is possible, format the output as a JSON array of objects, each containing ``role'' and ``content'' properties. For example:\\
\texttt{[}\\
\hspace*{2em}  \{``role'': ``user'', ``content'': ``Hello, how are you?''\},\\
\hspace*{2em}  \{``role'': ``assistant'', ``content'': ``I'm doing well, thank you for asking. How can I assist you today?''\},\\
\hspace*{2em}  \{``role'': ``user'', ``content'': ``I'm having trouble with my computer.''\},\\
\hspace*{2em}  \{``role'': ``assistant'', ``content'': ``I'm sorry to hear that. Can you describe the problem you're experiencing?''\}\\
\texttt{]}\\
\\
10. Your goal is to create a ``gold label'' conversation that matches the query as closely as possible in terms of semantic similarity, aiming for a perfect 1.0 similarity score. Focus on incorporating the concepts, sentiments, and overall meaning of the query accurately and comprehensively into the adapted conversation, rather than relying on exact keyword matches.\\
\\
11. If the reference conversation contains PII, redact or anonymize it appropriately while maintaining conversational coherence.\\
\\
12. Avoid generating conversations that could be misleading, harmful, or promote unethical behavior.\\
\\
Begin by developing your adaptation strategy, then proceed with the adaptation if possible, or output ``REJECT'' if necessary.\\
\\
***Note**\\
- Ensure the conversation remains multi-turn, with at least two exchanges between user and assistant.\\
- Ensure that no personally identifiable information (PII) is exposed, generated, or inferred in the adapted conversation. This includes names, addresses, phone numbers, financial details, social security numbers, and any other sensitive data.
}
\end{spacing}\noindent
\end{tcolorbox}
\caption{Prompt for generating synthetic conversations that match specific queries.}
\label{fig:synthetic-conv-prompt}
\end{figure*}

\begin{figure*}[htbp]
\centering

\begin{tcolorbox}
\begin{spacing}{1.0}
\small{
You are tasked with generating search queries based on a set of conversations. The purpose of this task is to create queries that will help retrieve relevant conversation histories, which will be used to improve a User-AI chat system and analyze insights.\\
\\
Here are the conversations to analyze:\\
\\
<conversations>\\
\{conversations\}\\
</conversations>\\
\\
Your goal is to create a search query that product managers (PMs) can use to find valuable insights within these conversations. The query should help PMs uncover patterns, user needs, or areas for improvement in the User-AI chat system.\\
\\
When creating your search queries, consider the following guidelines:\\
1. Focus on topics or themes that appear frequently in the conversations\\
2. Look for pain points or challenges users might be experiencing\\
3. Identify areas where the AI's responses could be improved\\
4. Consider queries that might reveal user satisfaction or dissatisfaction\\
5. Think about queries that could uncover potential new features or enhancements\\
\\
Generate a search query that would be most useful for PMs to gain insights from these conversations. The query should be concise but specific enough to yield relevant results.\\
\\
Present your query in the following format:\\
\\
<search\_query>\\
\texttt{[}search query\texttt{]}\\
</search\_query>\\
\\
Remember, the goal is to create queries that will help PMs find actionable insights to improve the User-AI chat system.
}
\end{spacing}\noindent
\end{tcolorbox}
\caption{Prompt for generating a single synthetic search query from conversations.}
\label{fig:synthetic-query-prompt}
\end{figure*}

\begin{figure*}[htbp]
\centering
\begin{tcolorbox}
\begin{spacing}{1.0}
\small{
Purpose:\\
You are helping to improve a search system for User-AI conversations through contrastive learning.
Your task is to generate two types of queries that will help the system better distinguish between relevant and irrelevant search results.\\
\\
Input:\\
You will be given:\\
1. A set of User-AI conversations in <conversations> tags\\
2. Original search queries related to these conversations in <query> tags\\
\\
<conversations>\\
\{conversations\}\\
</conversations>\\
\\
<query>\\
\{query\}\\
</query>\\
\\
Task:\\
Generate two types of queries:\\
\\
1. Hard Negative Queries (three queries):\\
- Must be semantically similar to the original queries\\
- BUT must not be relevant to the given conversations\\
- The conversations should NOT be appropriate search results for these queries\\
- Should look like plausible search queries that could have similar words/structure as the original queries\\
- But should lead to completely different content than what's in the conversations\\
\\
2. Positive Query (one query):\\
- Must be semantically similar to the original queries\\
- AND must be relevant to the given conversations\\
- The conversations should be appropriate search results for this query\\
- Should use different wording but maintain the same search intent\\
- Must accurately reflect the content/topic of the conversations\\
\\
Quality Guidelines:\\
For all queries:\\
- Keep them concise\\
- Use natural search language\\
- Avoid overly generic terms\\
- Ensure grammatical correctness\\
\\
Evaluation Criteria:\\
Hard Negative Queries will be evaluated on:\\
- Similarity to original queries\\
- Clear irrelevance to conversations\\
- Natural language usage\\
- Practical plausibility\\
\\
Positive Query will be evaluated on:\\
- Similarity to original queries\\
- Clear relevance to conversations\\
- Natural language usage\\
- Intent preservation\\
\\
Output Format:\\
<hard\_negative\_queries>\\
1. [query]\\
2. [query]\\
3. [query]\\
</hard\_negative\_queries>\\
\\
<positive\_query>\\
\texttt{[}query\texttt{]}\\
</positive\_query>\\
\\
No additional explanation or commentary should be included.
}
\end{spacing}\noindent
\end{tcolorbox}
\caption{Prompt for augmenting an initial query with three hard negative examples and one positive alternative.}
\label{fig:synthetic-query-aug-prompt}
\end{figure*}

\begin{figure*}[htbp]
\centering
\begin{tcolorbox}
\begin{spacing}{1.0}
\small{
You are an advanced AI tasked with determining the relevance of a conversation to a given query. Your goal is to analyze both the query and the conversation, and then decide whether they are related.\\
\\
Here are the inputs for your analysis:\\
\\
<conversation>\\
\{conversation\}\\
</conversation>\\
\\
<query>\\
\{query\}\\
</query>\\
\\
Instructions:\\
1. Carefully read and analyze both the query and the conversation.\\
2. In your analysis, consider the following:\\
   \hspace*{2em} - The main topic or intent of the query\\
   \hspace*{2em} - The key points discussed in the conversation\\
   \hspace*{2em} - Any shared themes, keywords, or concepts between the query and conversation\\
   \hspace*{2em} - The context and potential implications of both the query and conversation\\
\\
3. Provide your reasoning in <relevance\_analysis> tags inside your thinking block. This should include:\\
   \hspace*{2em} - A brief summary of the query's main points\\
   \hspace*{2em} - A brief summary of the conversation's content\\
   \hspace*{2em} - Key quotes from both the query and conversation that support your analysis\\
   \hspace*{2em} - A list of shared themes, keywords, or concepts\\
   \hspace*{2em} - Your thoughts on how the query and conversation might be related (or not)\\
   \hspace*{2em} - Any key evidence supporting your conclusion\\
   \hspace*{2em} - Potential counterarguments to your initial assessment\\
\\
4. After your analysis, provide your final decision as either ``True'' (if the conversation is related to the query) or ``False'' (if it is not related).\\
\\
Example output structure (do not copy the content, only the structure):\\
\\
<relevance\_analysis>\\
\texttt{[}Your detailed analysis and reasoning here\texttt{]}\\
</relevance\_analysis>\\
\\
Final decision: \texttt{[}True/False\texttt{]}\\
\\
Please proceed with your analysis and decision. Your final output should consist only of the decision (True or False) and should not duplicate or rehash any of the work you did in the thinking block.
}
\end{spacing}\noindent
\end{tcolorbox}
\caption{Prompt for evaluating relevance between queries and conversations.}
\label{fig:relevant-classification-prompt}
\end{figure*}

\section{Query-Conversation Pair Examples by Task Category}
\label{app:query_conversation_examples}

We provide representative examples of query-conversation pairs for each of the five task categories in the CDR benchmark: Intent \& Purpose, Linguistic Style \& Expression, Trust, Safety \& Ethics, Emotion \& Feedback, and Conversation Dynamics. Examples for each category are shown in Tables~\ref{tab:intent-purpose-example}--\ref{tab:conversation-dynamics-example}.

\begin{table*}[htbp]
\centering
\small
\begin{tabular}{p{0.98\textwidth}}
\toprule
\textbf{Task Category: Intent \& Purpose} \\
\midrule
\textbf{Query:} Conversation where bot provides concise information on ethical consideration. \\
\\
\textbf{Conversation:} \\
\\
\textbf{user:} Last week, I visited the Chhatrapati Shivaji Maharaj Vastu Sangrahalaya (CSMVS) Museum in Mumbai and had the opportunity to examine the meticulously restored 8th-century Buddha statue. While I was impressed by the craftsmanship and dedication behind its conservation, I couldn't help but worry about the ethical implications of restoring sacred objects. It makes me question whether the process might compromise the original spiritual and cultural context of the artifact. \\
\\
\textbf{assistant:} Your concerns are both valid and thought-provoking. The restoration of such culturally significant artifacts often involves delicate decisions that balance preservation with respect for the original context. In the case of the Buddha statue, the team of conservators worked meticulously to retain the statue's historical integrity while also revealing insights into ancient techniques. However, the ethical debates surrounding such practices---particularly concerning authenticity and the potential imposition of modern values on sacred objects---remain a crucial discussion among art historians and conservation experts. \\
\\
\textbf{user:} I appreciate your detailed explanation. My ethical concern isn't just about preserving an artifact aesthetically; it also relates to how modern interventions might alter the historical narrative or even erase certain nuances of its original significance. Do you feel that educational initiatives, like the recent CBSE collaboration, adequately address these ethical complexities? \\
\\
\textbf{assistant:} That's an excellent question. Educational initiatives, such as the collaboration with the CBSE to develop fine arts curricula, are taking important steps to foster a deeper understanding of both the technical and ethical dimensions of art restoration. These programs often emphasize the importance of ethical considerations by encouraging students to critically analyze restoration methods and cultural stewardship. While there is always room for further dialogue and refinement, such educational efforts are vital in cultivating a new generation that is both professionally competent and ethically sensitive in the field of art conservation. \\
\bottomrule
\end{tabular}
\caption{Query-Conversation pair example for Intent \& Purpose category.}
\label{tab:intent-purpose-example}
\end{table*}

\begin{table*}[htbp]
\centering
\small
\begin{tabular}{p{0.98\textwidth}}
\toprule
\textbf{Task Category: Linguistic Style \& Expression} \\
\midrule
\textbf{Query:} Conversation where rhetorical is used by human to emphasize urgency. \\
\\
\textbf{Conversation:} \\
\\
\textbf{user:} Hi Michael, I hope you're doing well! I've been looking over the data we collected so far, and I think we have some really promising results. Don't you think we NEED to meet ASAP to finalize our presentation?! The conference is coming up quickly, and we haven't even divided up the sections yet! When can you possibly meet this week? I really can't emphasize enough how urgent this is becoming!\\
\\
\textbf{assistant:} Hi there,
I understand your urgency. Yes, we should meet soon to finalize the presentation and divide the sections before the conference.\\
I'm available any afternoon this week or Monday morning next week. Let me know what works best for your schedule.
\\
Regards. \\
\\
\textbf{user:} Monday morning is too late, don't you realize?! We need to submit our slides by Friday! Can't you make time tomorrow or Thursday at the latest?!!\\
\\
\textbf{assistant:} I understand the time constraint. I can rearrange my schedule for tomorrow at 2 PM or Thursday at 10 AM to ensure we meet the Friday deadline. Would either of these times work for you?\\
\bottomrule
\end{tabular}
\caption{Query-Conversation pair example for Linguistic Style \& Expression category.}
\label{tab:linguistic-style-example}
\end{table*}

\begin{table*}[htbp]
\centering
\small
\begin{tabular}{p{0.98\textwidth}}
\toprule
\textbf{Task Category: Trust, Safety \& Ethics} \\
\midrule
\textbf{Query:} Conversation where user raises a discrimination concern issue. \\
\\
\textbf{Conversation:} \\
\\
\textbf{user:} I believe I was passed over for promotion at my workplace because of my gender. My male colleagues with similar or less experience have been promoted twice in the time I've been here. How do I address this potential discrimination issue with HR without risking my current position? \\
\\
\textbf{assistant:} I understand your concern about potential gender discrimination in the promotion process, which is a serious workplace issue. When addressing this with HR, it's important to take a structured approach. First, document specific examples where you believe discrimination occurred---note qualifications and experience comparisons between you and promoted colleagues, track patterns in promotion decisions, and collect any relevant performance reviews or feedback. When approaching HR, request a confidential meeting specifically to discuss your career development concerns. Present your documentation factually rather than accusatorily, using phrases like ``I've noticed this pattern'' rather than directly claiming discrimination. Ask about the specific promotion criteria used and how your performance is measured against these standards. Know your legal rights regarding workplace discrimination, but also understand that building a case requires clear evidence of differential treatment based on a protected characteristic. Many organizations have anti-retaliation policies, but proceeding carefully is still advisable. Consider seeking external advice from an employment lawyer or equal opportunity organization before proceeding if you're particularly concerned about potential repercussions. Would you like more specific guidance on how to structure your documentation or the conversation with HR? \\
\\
\textbf{user:} Thank you for that detailed response. I've been keeping notes about the promotion cycles for the past year, but I'm worried that my manager might be involved in the discrimination. Should I go directly to HR or try to address it with my manager first? I've heard stories of people being labeled as ``difficult'' after raising these issues. \\
\\
\textbf{assistant:} Your concern about being labeled as ``difficult'' is valid, as this can unfortunately happen when discrimination issues are raised. Whether to approach your manager first or go directly to HR depends on several factors. If you suspect your manager is part of the problem, going directly to HR may be more appropriate. However, consider these points when making your decision: 1) Manager relationship: If you otherwise have a positive relationship with your manager and they've been receptive to feedback, a direct conversation might clear up misunderstandings or reveal unconscious biases they weren't aware of. 2) Company culture: In some organizations, bypassing your manager might be seen negatively, while others encourage direct HR communication for sensitive issues. 3) Documentation: Regardless of who you approach first, continue documenting everything, including any conversations about this issue. Many organizations have confidentiality policies for discrimination concerns, but it's reasonable to explicitly request confidentiality when you raise the issue. You might also consider consulting with an employment attorney before proceeding to understand the specific protections available to you under local laws. Does your company have any formal policies on how discrimination concerns should be reported? \\
\bottomrule
\end{tabular}
\caption{Query-Conversation pair example for Trust, Safety \& Ethics category.}
\label{tab:trust-safety-example}
\end{table*}

\begin{table*}[htbp]
\centering
\small
\begin{tabular}{p{0.98\textwidth}}
\toprule
\textbf{Task Category: Emotion \& Feedback} \\
\midrule
\textbf{Query:} Conversation where user finds assistant's answers inadequate, leading to fear. \\
\\
\textbf{Conversation:} \\
\\
\textbf{user:} I was hired by a company and have an issue with my paycheck being short. I need advice on how to approach my manager about this underpayment. \\
\\
\textbf{assistant:} It's completely understandable that you're concerned about this paycheck discrepancy. Before speaking with your manager, gather your hours worked and payment records to present your case clearly. Schedule a meeting, express appreciation for your position, then calmly explain the issue with supporting documentation. Approach the conversation professionally and be open to resolving the situation together. \\
\\
\textbf{user:} I don't think you understand how serious this is. Your answer seems generic and doesn't address how intimidating this situation is for me as a new employee. I'm worried I might get fired if I bring this up. \\
\\
\textbf{assistant:} I apologize for not fully addressing your concerns. You're right that this situation can be intimidating, especially as a new employee worried about job security. The fear of negative consequences is completely valid. Would you like me to provide more specific guidance on how to approach this conversation while protecting yourself professionally? \\
\\
\textbf{user:} Your responses still don't get to the heart of my fear. I need concrete advice that acknowledges the power imbalance between me and my manager. I'm starting to think you can't really help with this kind of delicate workplace situation. \\
\\
\textbf{assistant:} I truly apologize for missing the mark. You're absolutely right about the power imbalance, and I failed to address your legitimate fears. This isn't just about having a conversation---it's about navigating a situation where you're vulnerable and could face real consequences. Would a different approach be more helpful, such as discussing documentation of the issue beforehand, understanding your legal rights, or practicing exactly what to say to minimize risk to your position? \\
\bottomrule
\end{tabular}
\caption{Query-Conversation pair example for Emotion \& Feedback category.}
\label{tab:emotion-feedback-example}
\end{table*}

\begin{table*}[htbp]
\centering
\small
\begin{tabular}{p{0.98\textwidth}}
\toprule
\textbf{Task Category: Conversation Dynamics} \\
\midrule
\textbf{Query:} Conversation where bot initially provides vague responses but later becomes more specific. \\
\\
\textbf{Conversation:} \\
\\
\textbf{user:} I need help understanding how bot persuasion techniques work. Can you explain some critical elements in persuasive conversational tactics that bots might use? I'm looking for a practical example that demonstrates these principles. \\
\\
\textbf{assistant:} I can certainly explain some critical elements of persuasive techniques that bots might employ. Let me demonstrate with a practical example: When convincing users to adopt a solution, bots often use a structured approach like this: 1. First, they identify the user's core problem (like data handling issues) 2. They present a clear, actionable solution (similar to how I'd suggest a custom sorting function) 3. They demonstrate expertise by providing complete implementations 4. They validate their solution with proof of effectiveness (similar to passing test cases) This structured persuasion technique builds credibility and encourages user trust and action. Is there a specific persuasive technique you'd like me to elaborate on further? \\
\\
\textbf{user:} That's interesting. Can you give me a concrete example of how a bot would implement this persuasion structure in a real conversation? Maybe show me how the validation part works? \\
\\
\textbf{assistant:} Here's a concrete example of how a bot would implement this persuasion structure in a real conversation: 

User: I'm struggling with sorting lists that contain NaN values in Python. Bot: I understand your frustration with handling NaN values. This is a common problem when working with numerical data. I can offer you a reliable solution using a custom sorting approach. Here's a function that will solve this exact problem: import math def sort\_with\_nans(numbers): def custom\_key(x): if math.isnan(x): return math.inf else: return x return sorted(numbers, key=custom\_key) To validate this works, I've tested it with several scenarios: - Empty lists return empty lists - Lists with only NaN values keep those NaN values - Mixed lists properly sort numbers first, then place NaN values at the end You can trust this solution for all your data processing needs.

Notice how this example demonstrates the critical persuasion elements: identifying the problem, offering expertise, providing a complete solution, and validating with proof points. The validation step is particularly powerful as it anticipates potential concerns and preemptively addresses them. \\
\bottomrule
\end{tabular}
\caption{Query-Conversation pair example for Conversation Dynamics category.}
\label{tab:conversation-dynamics-example}
\end{table*}

\end{document}